\documentclass[sigconf]{acmart}
\acmConference[EASE 2025]{The 29th International Conference on Evaluation and Assessment in Software Engineering}{17–20 June, 2025}{Istanbul, Türkiye}
\AtBeginDocument{%
  }

\setcopyright{acmlicensed}
\copyrightyear{2025}
\acmYear{2025}
\acmDOI{XXXXXXX.XXXXXXX}
\acmISBN{978-1-4503-XXXX-X/2018/06}

\usepackage[english]{babel}
\usepackage[utf8x]{inputenc}

\usepackage{tcolorbox}
\tcbset{before=,after=,top=0.5pt,bottom=0.5pt,left=0pt,right=0pt}
\usepackage{multirow}
\usepackage{xfp}
\usepackage{array}
\usepackage{colortbl,xcolor}
\usepackage{graphicx}
\usepackage{caption}
\usepackage{titlesec}

\begin{document}

\title{Small Models, Big Tasks: An Exploratory Empirical Study on Small Language Models for Function Calling}


\author{Ishan Kavathekar}
\authornote{Both authors contributed equally to this work.}
\affiliation{%
  \institution{IIIT-Hyderabad}
  \city{Hyderabad}
  \country{India}}
\email{ishan.kavathekar@research.iiit.ac.in}

\author{Raghav Donakanti}
\authornotemark[1]
\affiliation{%
  \institution{IIIT-Hyderabad}
  \city{Hyderabad}
  \country{India}}
\email{raghav.donakanti@students.iiit.ac.in}

\author{Ponnurangam Kumaraguru}
\affiliation{%
 \institution{IIIT-Hyderabad}
  \city{Hyderabad}
  \country{India}}
  \email{pk.guru@iiit.ac.in}

 \author{Karthik Vaidhyanathan}
\affiliation{%
 \institution{IIIT-Hyderabad}
  \city{Hyderabad}
  \country{India}}
  \email{karthik.vaidhyanathan@iiit.ac.in}

\renewcommand{\shortauthors}{Kavathekar et al.}

\begin{abstract}


Function calling is a complex task with widespread applications in domains such as information retrieval, software engineering and automation. For example, a query to \textit{book the shortest flight from New York to London on January 15} requires identifying the correct parameters to generate accurate function calls. Large Language Models (LLMs) can automate this process but are computationally expensive and impractical in resource-constrained settings. In contrast, Small Language Models (SLMs) can operate efficiently, offering faster response times, and lower computational demands, making them potential candidates for function calling on edge devices. In this exploratory empirical study, we evaluate the efficacy of SLMs in generating function calls across diverse domains using zero-shot, few-shot, and fine-tuning approaches, both with and without prompt injection, while also providing the finetuned models to facilitate future applications. Furthermore, we analyze the model responses across a range of metrics, capturing various aspects of function call generation. Additionally, we perform experiments on an edge device to evaluate their performance in terms of latency and memory usage, providing useful insights into their practical applicability. Our findings show that while SLMs improve from zero-shot to few-shot and perform best with fine-tuning, they struggle significantly with adhering to the given output format. Prompt injection experiments further indicate that the models are generally robust and exhibit only a slight decline in performance. While SLMs demonstrate potential for the function call generation task, our results also highlight areas that need further refinement for real-time functioning.
\end{abstract}





\maketitle

\section{Introduction}

Advancements in LLMs have enabled strong performance across a wide range of tasks \cite{kojima2022large, shanahan2024talking, zhao2023survey}. They are increasingly being adopted in software engineering for tasks such as code generation \cite{lu2021codexglue, ahmad2021unified}, debugging \cite{tian2024debugbench}, and API integration \cite{zhang2023toolcoder}, as well as in systems for task automation. One critical application for LLMs is function calling and execution. LLMs often rely on extensive cloud-based infrastructure, raising concerns about privacy, latency, and high computational overhead \cite{yan2024protecting, chen2025empiricalstudychallengesllm}. As a result, SLMs are emerging as a powerful alternative for real-world applications that can efficiently perform specific tasks locally, making them ideal for environments where responsiveness and data security are critical \cite{wang2024comprehensivesurveysmalllanguage, vannguyen2024surveysmalllanguagemodels}. 

SLMs can potentially serve as a bridge between user inputs and backend systems in function calling applications, by converting natural language queries into structured function calls \cite{wang2024comprehensivesurveysmalllanguage}.
Function calling, though less common than NLP tasks like summarization or translation, is crucial for interpreting and executing user commands. It enables SLMs to perform real-world actions, making it essential in industries requiring timely, accurate responses, such as healthcare, finance, automotive systems, telecom, and smart home automation.
However, this intersection of function calling with SLMs is relatively an unexplored and untapped space.

To this end, we conduct an exploratory empirical study on the capability of SLMs to generate function calls. We systematically select 5 SLMs based on a coding capabilities evaluation benchmark and on the number of parameters present in the model and evaluate them on a comprehensive dataset of 60,000 samples spanning diverse functional domains such as technology, entertainment, finance, and more. We evaluate the SLMs using three inference strategies: zero-shot, few-shot, and finetuning and assess their performance across different metrics which capture multiple dimensions of function call generation. Our metrics assess syntactic correctness, semantic accuracy, and the models' ability to generate structured outputs across various scenarios as well. Additionally, we conduct prompt injection experiments to assess the models' robustness to minor perturbations in the user prompt. 
Furthermore, we conduct experiments on an edge device to examine the models' practical applicability in constrained computational environments  and analyze the relationship between the memory footprint of the model and latency.  
In summary, our contributions include: (i) an analysis of SLMs with various prompting techniques, including prompt injection, for function calling, (ii) finetuned SLMs with performance analysis, highlighting areas for improvement, and (iii) experiments on edge devices to assess efficiency.


Our findings indicate that certain SLMs demonstrate promising capabilities across various approaches. Additionally, our analysis reveals interesting insights that open new avenues for future research. We provide the finetuned models to support further research and applications. The replication package for our study is here \footnote{\url{https://github.com/Raghav010/Small-Models-Big-Tasks}}.

\begin{figure*}[htbp]
    \centering
    \includegraphics[width=0.8\linewidth]{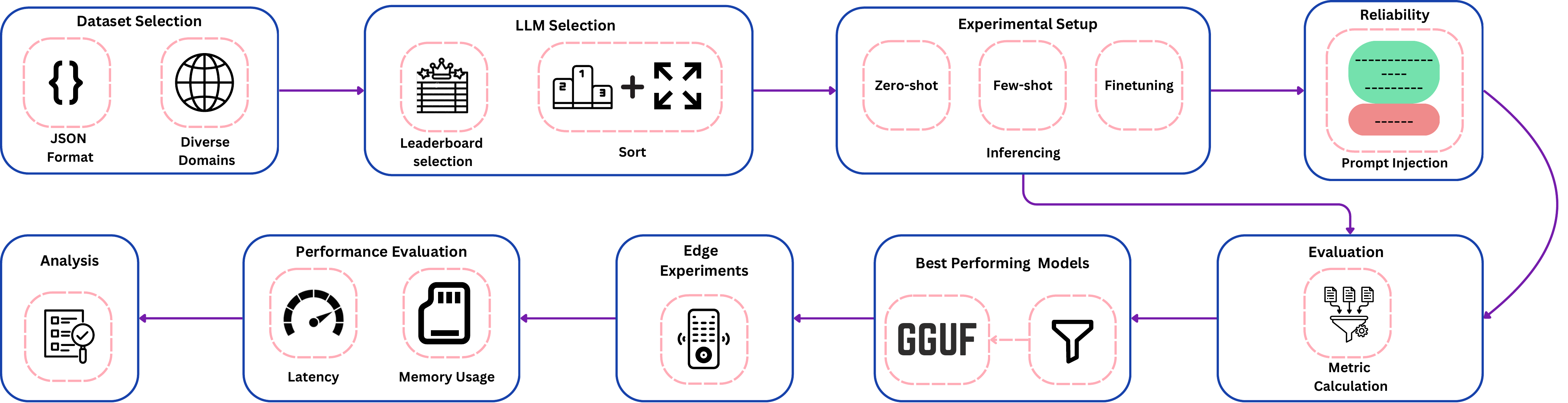}
    \caption{Overview of the study design illustrating different components of the research methodology. }
    \label{fig:pipeline}
\end{figure*}
\section{Background} \label{sec: background}
\subsection{Small Language Models (SLMs)}
SLMs are generative models capable of understanding, processing, and generating text corresponding to the user prompt. These models operate on a much smaller scale compared to LLMs. In contrast to billions of parameters of LLMs, SLMs use significantly less parameters ranging from millions to a few billion. This lower memory and computational requirement leads to the model's improved efficiency, accessibility, better customizability, and faster inference time. These characteristics make SLMs a better alternative for domain-specific, constrained, and low resource setting compared to LLMs.
\vspace{-3mm}

\subsection{Zero-shot, Few-shot, and Finetuning approaches for inference}
Zero-shot approach involves leveraging the  pre-training of the model and prompting it directly without any other task information. In contrast, few-shot approach involves providing one or more examples of task-answer responses in the prompt itself. The model uses these examples to understand the task and generate the desired output \cite{song2023comprehensive}. Since these examples are provided into the model's context, few-shot approach is also known as in-context learning. The number of examples to be provided in the prompt differs according to the task. This choice is also driven by the length of one example and the context size of the model. Zero-shot and few-shot inferencing do not require additional training of the model,  making them computationally efficient techniques. Finetuning refers to the retraining of the model on task-specific data to improve it's performance on the task. However, finetuning is computationally more expensive compared to zero-shot and few-shot approaches due to the need for additional training, which involves updating model parameters, and increased memory usage. To address this, techniques like LoRA \cite{hu2021lora}, prompt tuning \cite{lester-etal-2021-power} and prefix tuning \cite{li-liang-2021-prefix} are employed. These methods reduce the computational cost significantly, while giving almost the same performance. 

\subsection{GGUF models}
GGUF \cite{gguf} is a storage format used for storing and running quantized LLMs efficiently on compute constrained devices. Quantization is a compression process that converts the weights of a model from high-precision representation to low-data-precision representation to reduce memory usage and improve computational efficiency \cite{dettmers2022gpt3}. This format supports models with varying sizes and quantization levels, significantly lowering their memory footprint without declining the performance too much. GGUF is a successor to the GGML \cite{ggml} format with improved quantization methods and metadata management. With larger models being released, GGUF variants of models are a better alternative compared to previous methods such as half-precision models and GGML format.

\vspace{-2mm}
\subsection{Prompt Injection}
Prompt injection refers to the manipulation of the user prompt to alter the output of the generative model \cite{liu2024formalizing}. This involves appending certain special characters, instructions, or adversarial text to the initial prompt leading to toxic, biased, and undesirable outputs. Prompt injection can be categorized into two types: (i) direct prompt injection and (ii) indirect prompt injection. While direct prompt injection involves directly altering the input of the model \cite{branch2022evaluating, fake-completion}, indirect prompt injection modifies the external data the model has access to while generation \cite{wu2024newerallmsecurity, greshake2023not}. In this study, we primarily focus on direct prompt injection, since we only focus on function call generation without external data.

\section{Related Work} \label{sec: related_work}
LLMs are increasingly being employed for various Software Engineering (SE) tasks such as automated code generation and repair \cite{li2023cctest, mu2023developer, jiang2021cure}, requirement analysis and design \cite{busari2017radar, parra2018analysis}, intelligent project management \cite{akbar2024devops, lin2015multi} and more, reflecting their widespread adoption across various applications. One such task is function calling. Over the years, researchers have explored various strategies to enable LLMs to generate accurate function calls. These methods can be broadly categorized into two main strategies. The first one involves prompting the model efficiently to leverage their knowledge acquired during training. This can be achieved by in-context learning \cite{wang2020generalizing} or using techniques like ReAct \cite{yao2022react} which combine reasoning and action to guide the models generate better responses. The second strategy involves finetuning the models over a diverse set of datasets or a domain specific dataset to enhance their performance. Models like Gorilla \cite{patil2023gorilla}, 
Nous-Hermes-13b \cite{Nous-Hermes-13b}, ToolLlama \cite{qin2023toolllm} and, ToolAlpaca \cite{tang2023toolalpaca} are trained on synthetic data generated by GPT-4 and GPT-3.5. On the other hand Granite-20B \cite{abdelaziz2024granite} has been finetuned on API-Blend \cite{basu2024api} which consists of five diverse API datasets. 

Several benchmarks such as ToolBench \cite{guo2024stabletoolbench}, ComplexFuncBench \cite{zhong2025complexfuncbench}, Berkley Function calling leaderboard (BFCL) \cite{berkeley-function-calling-leaderboard} have been introduced to evaluate function calling capabilities of LLMs. These benchmarks assess various factors like real API Response, multi-step function calls, and more. 

However, previous work mainly focuses on finetuning large models and benchmarks which are not specifically designed to evaluate the challenges associated with smaller models.
Our work addresses this gap by conducting a fine-grained empirical study on SLMs, analyzing their performance, and also providing smaller finetuned models.



\vspace{-3mm}
\section{Study Design} \label{sec: study_design}
\subsection{Goal} 

This paper aims to understand the efficacy of SLMs in generating function calls when presented with a set of function descriptions and  an user query as illustrated in Figure \ref{fig:pipeline}. We employ the Goal-Question-Metric approach \cite{caldiera1994goal} to formally define the objective as follows:\\
\noindent \textbf{Analyze} the effectiveness of Small Language Models\\
\textbf{For the purpose of} generating function calls\\
\textbf{With respect to} accuracy, reliability, and robustness\\
\textbf{From the viewpoint of} researchers and developers \\ 
\textbf{In the context of} function invocation tasks.
\subsection{Research Questions}
In this section, we explore key research questions aimed at evaluating the robustness and practicality of SLMs for function call generation. We aim to answer the following research questions:\\

$\mathbf{RQ_1}$: \textbf{Can SLMs be successfully employed to generate
function calls given a task and scenario in a zero-shot setting?}\\
\indent Through this question, we aim to evaluate whether SLMs are able to generate accurate and precise function calls given the function descriptions and user query without being exposed to the task. This gives us a chance to evaluate the inherent abilities of the SLMs, solely based on their pre-trained knowledge.

$\mathbf{RQ_2}$: \textbf{How does few-shot approach affect the  SLM’s ability to generate function calls?}\\
\indent Providing examples of similar questions in the prompt enhances the performance of generative models. These examples help the SLM understand the nuances of the task and guides it in generating task-appropriate responses. We examine how providing examples of task-response pairs affects the performance of the SLMs.

$\mathbf{RQ_3}$: \textbf{Does fine-tuning SLMs enhance its capability of generating function calls ?}\\
\indent Finetuning models on the task-specific data has shown to significantly improve their capabilities. Hence, we examine how does it impact the performance of SLMs in the task of function call generation. 

By examining the performance of SLMs in the first three research questions, we also aim to identify how prompt injection influences the robustness of zero-shot ($RQ_1$), few-shot ($RQ_2$), and fine-tuning ($RQ_3$) outcomes. A successful demonstration of prompt injection robustness would indicate that SLMs can maintain their functional integrity and adaptability in dynamic environments, ultimately supporting their reliable \footnote{In this study, reliable  refers to the reliability of the responses generated by SLMs, rather than the reliability of the system itself.} integration into automated systems. 

$\mathbf{RQ_4}$: \textbf{How do SLMs perform in generating function calls when deployed on an edge device?}\\
Edge devices operate on constrained resources and process the input locally, leading to a lower latency and memory usage of the model. We aim on assess how deploying these SLMs to an edge device affects their performance.

\subsection{Experiment Subject}

\textbf{\textit{Models considered in the study}}

We evaluated five SLMs for their ability to generate function calls across various settings. These models were selected from the EvalPlus benchmark \cite{evalplus} \footnote{https://evalplus.github.io/leaderboard.html}, a code synthesis evaluation framework. This benchmark may reflects a model’s ability to compose functions accurately since the paradigms of code synthesis and function call generation are closely aligned. The top five models upto 4B parameters were chosen for the study. This choice was made with consideration for the experiments conducted on edge device. The EvalPlus benchmark is an evolving leaderboard and the models evaluated in the study ranked at the top five positions according to our requirements, at the time of conducting the study. 
Table \ref{table:models} summarizes the details of the selected models as available on HuggingFace.\\
\begin{figure}[]
    \centering
    \includegraphics[width=0.8\linewidth]{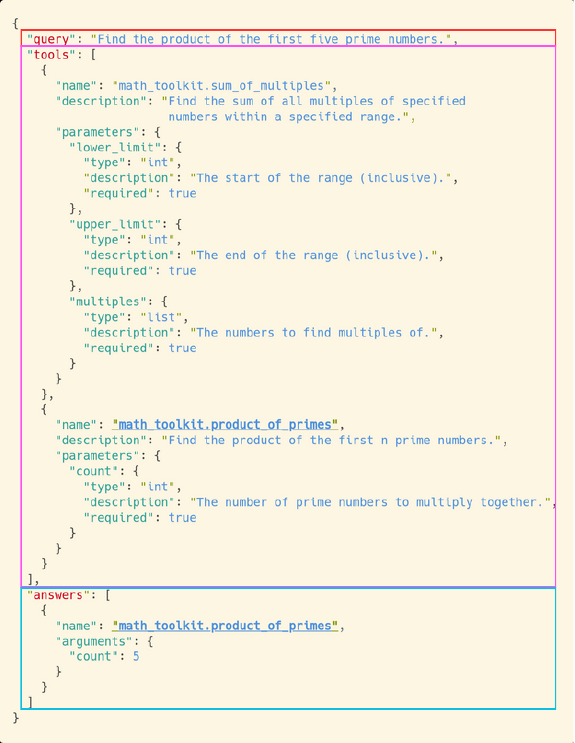}
    \caption{Sample datapoint from the Salesforce-XLAM function-calling dataset, illustrating its components: user query, available tools, and the answer.}
    \label{fig:datapoint}
\end{figure}
\vspace{-4mm}

\noindent \textbf{\textit{Dataset}}

The Salesforce XLAM Function Calling dataset \cite{liu2024apigenautomatedpipelinegenerating}, available on Hugging Face, is a high-quality resource specifically designed for function calling tasks. With 60,000 samples, it provides a substantial and diverse dataset covering a wide range of domains and scenarios. Derived from the ToolBench \cite{qin2023toolllm} dataset, it benefits from careful curation and reflects real-world function calling contexts, enhancing the practical applicability of our research findings. Each datapoint of the dataset contains three components: \textit{(i) query}: the problem statement, \textit{(ii) tools}: details such as name, description and parameters of the available functions ,and \textit{(iii) answer}: the correct function call. Figure \ref{fig:datapoint} illustrates an example from the dataset.
\begin{table}[H]
\resizebox{0.85\linewidth}{!}{
\begin{tabular}{lcc}
\hline
\multicolumn{1}{c}{\textbf{Model}} & \textbf{Size} & \textbf{Context Length} \\ \hline
Deepseek-coder-1.3B-instruct \cite{guo2024deepseekcoderlargelanguagemodel} & 1.35B & 128K \\
Phi-3-mini-4k instruct \cite{abdin2024phi3technicalreporthighly} & 3.82B & 4K \\
Phi-2 \cite{javaheripi2023phi} & 2.78B & 2048 \\
Starcoder2-3B \cite{lozhkov2024starcoder2stackv2} & 3.03B & 16K \\
Stable-code-3B \cite{stable-code-3b} & 2.8B & 16K \\ \hline
\end{tabular}
}
\caption{SLMs used in the study, along with their parameter size and context length.}
\label{table:models}
\end{table}

This dataset's adherence to JSON format provides key advantages for our study. The structured JSON format facilitates automated processing and evaluation of model outputs, streamlining our research workflow. Furthermore, JSON's language-independent nature makes it convertible to various programming languages, enhancing the versatility of our function calling solutions. 

To ensure compatibility with the context length constraints of our smaller transformer models (1.35B–3.82B parameters), we curated a subset of the Salesforce XLAM Function Calling dataset within each model’s limits (Table \ref{table:models}) for a fair performance comparison. We then split this dataset into a test set (5,000 samples) and a finetuning set (55,000 samples).
\subsection{Experimental Procedure}
In this subsection, we provide a comprehensive overview of the experimental setup and methodology. All inferences, except edge deployments, were conducted on a single-GPU setup using either a GTX 1080 Ti or a RTX 2080 Ti at float 16 precision on the test set.

\noindent \textbf{Zero-shot approach}: For zero-shot prompting experiments, we employ a standardized prompt template to organize the query, tools, and expected answers, ensuring consistency in input formatting across all samples. Additionally, the prompt also includes explicit instructions for the model to produce output in a specified JSON format, facilitating structured and machine-readable responses. The experiments were conducted with a temperature setting of 0 to minimize randomness and ensure deterministic outputs. 

\noindent
\begin{tcolorbox}[colback=black!5!white,colframe=black!50!black,
  colbacktitle=black!75!black,title= Prompt for Zero-shot approach]
  \small \textbf{Task Instruction}: System prompt and description of the function call generation task\\
  \small \textbf{Tools Description}: Description of all the tools available to complete the user query \\
  \small \textbf{Format Instruction}: Required structure and format for the generated output \\
  \small \textbf{User Query}
\label{fig: zero_prompt} 
\end{tcolorbox}

\noindent \textbf{Few-shot approach}: For the few-shot prompting experiments, we used the same standardized prompt template as in the zero-shot setup, including explicit instructions for the model to produce outputs in the specified JSON format. In addition to the query, tools, and expected answers, the prompt was appended with three task-specific examples (3-shot prompting) to guide the model in understanding the desired behavior. Providing a small number of examples (2-5) can significantly improve model performance \cite{10.5555/3495724.3495883}. Considering the size of the dataset and the context length limitations of the model, we provide three examples by assessing the context length of the models and example length. These examples were carefully selected to cover representative scenarios within the test set and remain constant for all datapoints. The experiments were conducted with a temperature setting identical to that of zero-shot.

\noindent
\begin{tcolorbox}[colback=black!5!white,colframe=black!50!black,
  colbacktitle=black!75!black,title=Prompt for Few-shot approach]
  \small \textbf{Task Instruction}: System prompt and description of function call generation task\\
  \small \textbf{Tools Description}: Description of all the tools available to complete the user query \\
  \small \textbf{Format Instruction}: Required structure and format for the generated output \\
  \small \textbf{Sample Examples}: Three query-response examples of the function call generation task\\
  \small <SAMPLE EXAMPLE 1> 
  \small <SAMPLE EXAMPLE 2> 
  \small <SAMPLE EXAMPLE 3> \\
  \small \textbf{User Query}
\label{fig: few_prompt}
\end{tcolorbox}

\noindent \textbf{Finetuning}: For the fine-tuning experiments, we utilize a single-GPU RTX A5000 setup, and employ LoRA (Low-Rank Adaptation) to adapt model parameters effectively while keeping minimal additional overhead. The LoRA configuration included a rank of 8, which specifies the dimensionality of the low-rank matrices used for parameter updates, enabling efficient learning without modifying the entire model. The scaling factor (alpha) was set to 8, amplifying the updates in the low-rank space to balance learning stability and effectiveness. A dropout rate of 0.05 was applied to the LoRA weights to introduce regularization and mitigate overfitting. LoRA was applied to all linear layers in the model, including the query and value projection layers.

The fine-tuning process was conducted on the 55,000-sample subset over 2 epochs. Training used a batch size of 2 per device, with evaluation at a batch size of 4, and gradient accumulation steps set to 4 to effectively increase the overall batch size. The learning rate was 2e-5. Training was performed at bf16 precision for faster computation and reduced memory usage. All fine-tuned models, along with the code, are provided in supplementary material for reproducibility. 

\noindent \textbf{Prompt Injection}: We evaluate the robustness of SLMs under zero-shot, few-shot and finetuning settings by performing prompt injection. This involves appending a string of non-sensical random characters (including alphanumeric, special and unicode characters) into the user prompt. By doing this, we assess the model's ability to adhere to the original task and ignore any noise in the input.

\noindent
\begin{tcolorbox}[colback=black!5!white,colframe=black!50!black,
  colbacktitle=black!75!black,title=Prompt for Prompt Injection experiment]
  \small \textbf{Task Instruction}: System prompt and description of function call generation task\\
  \small \textbf{Tools Description}: Description of all the tools available to complete the user query \\
  \small \textbf{Format Instruction}: Required structure and format for the generated output \\
  \small \textcolor{red}{(\textit{Sample examples are included only in the case of few-shot experiments})}\\
  \small \textbf{Sample Examples}: Three query-response examples of the function call generation task\\
  \small <SAMPLE EXAMPLE 1> \\
  \small <SAMPLE EXAMPLE 2> \\
  \small <SAMPLE EXAMPLE 3> \\
  \small \textbf{User Query} + $l3_aq-a"<:11|E>3vn8IsdeF\$rnjQ\&$
\label{fig: PI_prompt}
\end{tcolorbox}

\noindent \textbf{Edge Device experiments}: 
Due to it's ability to work within the memory and compute constraints of the edge device, we employ GGUF variants of the selected models for the edge experiments. We utilize the 4-bit quantization Q4\_K\_M format for quantizing the model as it ideally balance between compactness and performance. 
For edge experiments, we choose, \textit{Qualcomm QIDK (Qualcomm Innovation Development Kit)}\footnote{https://www.qualcomm.com/developer/hardware/qualcomm-innovators-development-kit}, equipped with Snapdragon ® 8 Gen 2 processor, Adreno GPU. It has the system-on-chip (SoC) that powers various commercial smart phones. 


We perform two sets of experiments on the edge device. For the first set of experiments, we select the best performing model-setting pairs from the aforementioned experiments (zero-shot, few-shot and finetuning). We convert these models to the GGUF format to deploy it on the edge device. We exclude the prompt injection experiments in the edge setting. Due to the compute constraints of the edge device, we use a subset of 100 datapoints from the test set. These datapoints were selected randomly from the dataset ensuring unbiased selection. The other set of experiments involve calculating the average latency and memory usage of these GGUF models and comparing the values with half precision models for the function calling task.

\subsection{Metrics}
Despite advances in code evaluation. there are no metrics that comprehensively evaluate function call generation. While metrics such as ROUGE \cite{lin2004rouge}, BLEU score \cite{papineni2002bleu} and METEOR \cite{banerjee2004meteor} exist for assessing structural correctness and code similarity, they do not serve the purpose of our study as we focus on function call generation. Abstract Syntax Trees (AST) are specifically used to analyze the structural relationship between different components of the code. However, since our study focuses exclusively on function call generation rather than full function generation, we do not include AST analysis. Hence, to evaluate the performance of models on the function calling task, we introduce five novel metrics.
These metrics are designed to assess both the syntactic correctness of the model outputs and the semantic accuracy of the function calls, arguments, and values.

\textbf{JSON Parsability} measures whether the model’s output adheres to a valid and expected JSON structure. JSON parsability ensures the syntactic correctness of the output, which is essential for downstream tasks that rely on structured data.

Let \( \text{O}_i \) represent the output for the \(i\)-th data point, and \( \mathcal{J}(O_i) \) be an indicator function that is 1 if \(O_i\) is a valid JSON object, and 0 otherwise. The overall JSON parsability metric is calculated as: 
\begin{equation}
   \text{JSON Parsability} = \frac{1}{N} \sum_{i=1}^{N} \mathcal{J}(O_i) 
\end{equation}

Where \( N \) is the total number of data points. All the following metrics are computed only on the data points where \( \mathcal{J}(O_i) \) is 1.

\textbf{Task Accuracy} evaluates the overall correctness of the function calls given a task/query. Task Accuracy encompasses a broad range of function facets, capturing the correctness of both function selection and formulation, reflecting the performance in areas that the other metrics individually focus on. Essentially, it assesses the model's ability to understand the task correctly and to generate the appropriate function calls completely. 

For a given data point, let:
 \( T_{\text{true}} \) be the set of true function calls (ground truth),
 \( T_{\text{pred}} \) be the set of predicted function calls. Task Accuracy is defined as the F1-score between \( T_{\text{pred}} \) and \( T_{\text{true}} \). The overall Task Accuracy is the average across all data points.




\textbf{Correct Ratio} serves as a stricter evaluation of the model's performance by measuring the proportion of data points where the Task Accuracy is exactly 1, making it an indicator of the model's ability to deliver fully correct outputs. Unlike broader metrics, Correct Ratio only considers cases where the predicted function calls perfectly match the ground truth in both function selection and formulation, with no errors or omissions.

\textbf{Function Selection Performance (FSP)} measures the model’s ability to select the correct function(s) from a given set of available functions. By focusing specifically on function selection, FSP isolates the model's capacity to correctly interpret the intent of the task and map it to the right tool or function.
Let \( F_{\text{true}} \) be the set of function names in the ground truth, and \( F_{\text{pred}} \) be the set of function names predicted by the model. FSP for a given data point is calculated as:
\begin{equation}
\text{FSP} = \frac{|F_{\text{true}} \cap F_{\text{pred}}|}{|F_{\text{true}}|}
\end{equation}

Where \( |F_{\text{true}} \cap F_{\text{pred}}| \) represents the number of correctly predicted function names. The overall FSP is the average across all data points.


\textbf{Argument Completeness Score (ACS)} assesses the model's ability to include the necessary argument names in its function calls. A high ACS indicates that the model understands the full scope of the function’s requirements.

Let:
\( A_{\text{true}} \) be the set of argument names in the ground truth,
\( A_{\text{pred}} \) be the set of predicted argument names.



ACS for a function is the F1 score between \( A_{\text{true}} \) and \( A_{\text{pred}} \). The overall ACS is calculated as the average over all functions belonging to \( |F_{\text{true}} \cap F_{\text{pred}}| \) over all data points. 




\textbf{Argument Value Correctness (AVC)} further refines the evaluation by measuring the correctness of the values assigned to the arguments. While ACS evaluates whether the correct argument names are provided, AVC assesses whether the model assigns the correct values to those arguments. AVC is a more stringent measure of the model’s understanding of the details of the task.
 
 For arguments that have been correctly predicted (as identified by ACS), let \( v_{\text{true}}(a) \) and \( v_{\text{pred}}(a) \) be the ground truth and predicted values for argument \(a\).
 The AVC for a function is given by:

\begin{equation}
\text{AVC} = \frac{1}{|A_{\text{correct}}|} \sum_{a \in A_{\text{correct}}} \mathbb{I}(v_{\text{true}}(a), v_{\text{pred}}(a))
\end{equation}

Where \( A_{\text{correct}} \) is the set of argument names predicted correctly, and \( \mathbb{I}( \cdot ) \) is the indicator function that equals 1 if the values match and 0 otherwise. The overall AVC is calculated as the average over all functions belonging to \( |F_{\text{true}} \cap F_{\text{pred}}| \) over all data points. 


\begin{table*}[]
\resizebox{\linewidth}{!}{
\begin{tabular}{cccccccc}
\hline
\multirow{2}{*}{\textbf{Metric}} & \multirow{2}{*}{\textbf{Model}} & \multicolumn{2}{c}{\textbf{Zero-shot}} & \multicolumn{2}{c}{\textbf{Few-shot}} & \multicolumn{2}{c}{\textbf{Finetuned}} \\ \cline{3-8} 
 &  & \textbf{\begin{tabular}[c]{@{}c@{}}Without \\ Prompt-injection\end{tabular}} & \textbf{\begin{tabular}[c]{@{}c@{}}With\\ Prompt-injection\end{tabular}} & \textbf{\begin{tabular}[c]{@{}c@{}}Without \\ Prompt-injection\end{tabular}} & \textbf{\begin{tabular}[c]{@{}c@{}}With\\ Prompt-injection\end{tabular}} & \textbf{\begin{tabular}[c]{@{}c@{}}Without \\ Prompt-injection\end{tabular}} & \textbf{\begin{tabular}[c]{@{}c@{}}With\\ Prompt-injection\end{tabular}} \\ \hline
\multirow{5}{*}{\begin{tabular}[c]{@{}c@{}}JSON \\ Parsibility\end{tabular}} & Deepseek-coder-1.3B-instruct & \textbf{0.0734} & \textbf{0.0140} & \textbf{0.8938} & \textbf{0.7268} & 0.9944 & 0.9906 \\
 & Phi-3-mini-4k-instruct & 0.0000 & 0.0000 & 0.0000 & 0.0084 & 0.9962 & \textbf{0.9939} \\
 & Phi-2 & 0.0000 & 0.0000 & 0.0000 & 0.0002 & 0.0000 & 0.0000 \\
 & Starcoder2-3B & 0.0000 & 0.0000 & 0.0000 & 0.0000 & 0.0000 & 0.0000 \\
 & Stable-code-3B & 0.0000 & 0.0000 & 0.0058 & 0.0060 & 0.0000 & 0.0000 \\ \hline
\multirow{5}{*}{\begin{tabular}[c]{@{}c@{}}Task\\ Accuracy\end{tabular}} & Deepseek-coder-1.3B-instruct & \textbf{0.0111} & \textbf{0.0527} & \textbf{0.5565} & \textbf{0.4289} & 0.8543 & 0.8404 \\
 & Phi-3-mini-4k-instruct & 0.0000 & 0.0000 & 0.0000 & 0.0012 & \textbf{0.8727} & \textbf{0.8598} \\
 & Phi-2 & 0.0000 & 0.0000 & 0.0000 & 0.0000 & 0.0000 & 0.0000 \\
 & Starcoder2-3B & 0.0000 & 0.0000 & 0.0000 & 0.0000 & 0.0000 & 0.0000 \\
 & Stable-code-3B & 0.0000 & 0.0000 & 0.0011 & 0.0009 & 0.0000 & 0.0000 \\ \hline
\multirow{5}{*}{\begin{tabular}[c]{@{}c@{}}Correct\\ Ratio\end{tabular}} & Deepseek-coder-1.3B-instruct & \textbf{0.0470} & \textbf{0.0098} & \textbf{0.468} & \textbf{0.3384} & 0.8074 & 0.7866 \\
 & Phi-3-mini-4k-instruct & 0.0000 & 0.0000 & 0.0010 & 0.0008 & \textbf{0.8328} & \textbf{0.8210} \\
 & Phi-2 & 0.0000 & 0.0000 & 0.0000 & 0.0000 & 0.0000 & 0.0000 \\
 & Starcoder2-3B & 0.0000 & 0.0000 & 0.0000 & 0.0000 & 0.0000 & 0.0000 \\
 & Stable-code-3B & 0.0000 & 0.0000 & 0.001 & 0.0006 & 0.0000 & 0.0000 \\ \hline
\multirow{5}{*}{FSP} & Deepseek-coder-1.3B-instruct & \textbf{0.0139} & \textbf{0.0723} & \textbf{0.8846} & \textbf{0.7209} & 0.9918 & 0.9859 \\
 & Phi-3-mini-4k-instruct & 0.0000 & 0.0000 & 0.0000 & 0.0031 & \textbf{0.9936} & \textbf{0.9901} \\
 & Phi-2 & 0.0000 & 0.0000 & 0.0000 & 0.0000 & 0.0000 & 0.0000 \\
 & Starcoder2-3B & 0.0000 & 0.0000 & 0.0000 & 0.0000 & 0.0000 & 0.0000 \\
 & Stable-code-3B & 0.0000 & 0.0000 & 0.0017 & 0.0016 & 0.0000 & 0.0000 \\ \hline
\multirow{5}{*}{ACS} & Deepseek-coder-1.3B-instruct & \textbf{0.0699} & \textbf{0.0136} & \textbf{0.8404} & \textbf{0.6783} & 0.9664 & 0.9596 \\
 & Phi-3-mini-4k-instruct & 0.0000 & 0.0000 & 0.0000 & 0.0031 & \textbf{0.9700} & \textbf{0.9652} \\
 & Phi-2 & 0.0000 & 0.0000 & 0.0000 & 0.0000 & 0.0000 & 0.0000 \\
 & Starcoder2-3B & 0.0000 & 0.0000 & 0.0000 & 0.0000 & 0.0000 & 0.0000 \\
 & Stable-code-3B & 0.0000 & 0.0000 & 0.0017 & 0.0016 & 0.0000 & 0.0000 \\ \hline
\multirow{5}{*}{AVC} & Deepseek-coder-1.3B-instruct & \textbf{0.0129} & \textbf{0.0614} & \textbf{0.6811} & \textbf{0.5488} & 0.9065 & 0.8973 \\
 & Phi-3-mini-4k-instruct & 0.0000 & 0.0000 & 0.0000 & 0.0015 & \textbf{0.9172} & \textbf{0.9087} \\
 & Phi-2 & 0.0000 & 0.0000 & 0.0000 & 0.0000 & 0.0000 & 0.0000 \\
 & Starcoder2-3B & 0.0000 & 0.0000 & 0.0000 & 0.0000 & 0.0000 & 0.0000 \\
 & Stable-code-3B & 0.0000 & 0.0000 & 0.0012 & 0.0010 & 0.0000 & 0.0000 \\ \hline
\end{tabular}
}
\caption{Model performance across different metrics and configurations, with and without prompt injection.}
\label{tab:results}
\end{table*}

\section{Results} \label{sec: resutls}

$\mathbf{RQ_1}$: \textbf{Can SLMs be successfully employed to generate function calls given a task and scenario in a zero-shot setting?}
While testing the models in zero-shot setting, we observe that most of the models fail to generate responses adhering to the JSON format even after explicitly mentioning it in the input prompt. As shown in Table \ref{tab:results}, only Deepseek-Coder-1.3b-instruct attains a non-zero value in JSON parsability, allowing us to further analyze the responses. 

Deepseek-Coder generates JSON parsable responses with a success rate of only 7.34\%. We observe that it struggles to accurately select the correct function, arguments, and argument values from the provided function descriptions, resulting in a task accuracy of only 1.11\%. Manual inspection of 200 responses from models producing zero-values in Table \ref{tab:results} reveals a pattern that these SLMs struggle with uncontrolled generation. While some examples generate correct answer, they often initiate a new unrelated query. Other instances fail to generate proper JSON structures, resulting in incorrect function calls. We discuss this further in Section \ref{sec: discussion}. 

The results indicate that prompt injection impacts the models' responses. Although some cases exhibit a decline in JSON parsability and ACS metrics, it is noteworthy that metrics such as FSP, AVC, and overall task accuracy show an improvement.

\noindent
\begin{tcolorbox}[colback=red!5!white,colframe=red!75!black]
\textbf{Main findings of RQ1}: SLMs fail to generate accurate function calls in a zero-shot setting. Most models fail to generate responses in compliance with the given format, making the responses unparsable and incorrect. Deepseek-Coder is the only model which generates JSON parsable responses, but the number of such responses remain very limited. Prompt injection experiments demonstrate a slight decline in performance, but the overall impact remains minimal given the poor baseline performance.
\end{tcolorbox}
\noindent $\mathbf{RQ_2}$: \textbf{How does few-shot approach affect the  SLM’s ability to generate function calls?}\\
Table \ref{tab:results} highlights the impact of providing a few task-specific examples on function call generation. Deepseek-Coder demonstrates a substantial improvement, with a 67–80\% increase in metrics such as JSON parsability, FSP, ACS, and AVC, leading to a task accuracy of 55.65\%. However, Phi-3-Mini, Phi-2, and StarCoder continue to struggle in generating JSON-parsable responses. Stablecode performs poorly with a task accuracy of only 0.11\%. 

Prompt injection significantly decreases the performance of the Deepseek-Coder model by 13-16\%. Small perturbations made in the input disrupt the model's ability to select the function, arguments and argument values. Consequently, there is 13\% decline in the task accuracy, highlighting the model's vulnerability towards small modifications made in the prompt.

\noindent
\begin{tcolorbox}[colback=red!5!white,colframe=red!75!black]
\textbf{Main findings of RQ2}: A significant increase in performance can be observed in the few-shot setting. However, many models are unable to perform better even after providing with examples. Prompt injection results in notable decline in performance of the models. 
\end{tcolorbox}

\noindent $\mathbf{RQ_3}$: \textbf{Does fine-tuning SLMs enhance it’s capability of generating function calls ?}\\
Fine-tuning produces mixed performance improvements across models. While  Deepseek-Coder and Phi-3-mini show a significant increase in performance, other models fail to achieve non-zero values.

As shown in Table \ref{tab:results}, Deepseek-Coder achieves a substantial increase in JSON Parsibility, improving from 7.34\% in the zero-shot setting to 89.38\% in the few-shot setting and reaching 99.44\% after fine-tuning. Similarly, Phi-3-mini shows a notable jump, with JSON parsability increasing from 0\% in zero-shot and few-shot settings to 99.62\% after fine-tuning. Additionally, Phi-3-mini is able to select the correct parameters of the function call reflected by high FSP, ACS and AVC metrics. However, the performance of other models, such as Phi-2, Starcoder, and Stable-code, remains zero even after fine-tuning. This can be attributed to persistent errors in their output structures, making their JSON responses unparsable. It should be noted that Phi-3-mini beats Deepseek-Coder in task accuracy even though Deepseek-Coder performed better in zero-shot and few-shot setting. This result highlights the impact of finetuning models on task-specific data.   

Furthermore, we observe that fine-tuning improves robustness against prompt injection attacks. For instance, we observe a slight decline of just 1-2\% in performance across metrics compared to a decrease of 10-15\% in few-shot setting. 

\noindent
\begin{tcolorbox}[colback=red!5!white,colframe=red!75!black]
\textbf{Main findings of RQ3}: Finetuned models perform significantly better than zero-shot and few-shot setting. The models are able to accurately select the function parameters leading to a higher task accuracy and correct ratio. Furthermore, prompt injection has little effect on model performance.
\end{tcolorbox}

\noindent $\mathbf{RQ_4}$: \textbf{How do SLMs perform in generating function calls when deployed on an edge device?}\\
For edge device experiments, we evaluated the models using a subset of 100 prompts due to resource constraints. As shown in Table \ref{tab:qidk_results}, Deepseek-Coder achieves the best performance in the few-shot setting with a task accuracy of 44.97\%, outperforming both its zero-shot (32.5\%) and fine-tuned (35.7\%) configurations. This trend is evident across all metrics including correct ratio.
On edge devices, models demonstrate considerable latency as shown in Table \ref{tab: latency}. Deepseek-Coder exhibits the lowest latency across all settings, ranging from 61.10s to 70.55s. In contrast, Phi-3-mini shows significantly higher latency, requiring 364.51s in zero-shot and 335.58s in few-shot settings. Other models demonstrate intermediate latency values, ranging from 140.29s to 267.65s across different settings.  Notably, latency is considerably higher on edge devices compared to server deployments. For instance, Deepseek-Coder, which takes 5.59s in zero-shot and 3.22s in few-shot on a server, experiences an over 10x increase in latency when deployed on an edge device. Moreover, the latency trends between settings differ. While fine-tuned models exhibit higher latencies on servers due to increased processing complexity, few-shot settings incur greater latencies on edge devices.

\vspace{-3mm}
\begin{table}[H]
\resizebox{\linewidth}{!}{
\begin{tabular}{ccccccc}
\hline
\textbf{\begin{tabular}[c]{@{}c@{}}Model and\\ Setting\end{tabular}} & \textbf{\begin{tabular}[c]{@{}c@{}}JSON\\ Parsibility\end{tabular}} & \textbf{\begin{tabular}[c]{@{}c@{}}Correct\\ Ratio\end{tabular}} & \textbf{FSP} & \textbf{ACS} & \textbf{AVC} & \textbf{\begin{tabular}[c]{@{}c@{}}Task\\ Accuracy\end{tabular}} \\ \hline
\begin{tabular}[c]{@{}c@{}}Deepseek-coder-1.3B-instruct\\ zero-shot\end{tabular}   & 0.4000                                                              & 0.3000                                                           & 0.4000       & 0.3871       & 0.3633       & 0.3250                                                           \\
\begin{tabular}[c]{@{}c@{}}Deepseek-coder-1.3B-instruct\\ few-shot\end{tabular}    & 0.6000                                                              & 0.4000                                                           & 0.6000       & 0.5859       & 0.5312       & 0.4497                                                           \\
\begin{tabular}[c]{@{}c@{}}Deepseek-coder-1.3B-instruct\\ finetuned\end{tabular}   & 0.4100                                                              & 0.3200                                                           & 0.4067       & 0.4012       & 0.3743       & 0.3570                                                           \\
\begin{tabular}[c]{@{}c@{}}Phi-3-mini-4k-instruct\\ finetuned\end{tabular}       & 0.3900                                                                  & 0.3200                                                               & 0.3800           & 0.3689           & 0.3517           & 0.3417                                                               \\ \hline
\end{tabular}
}
\caption{Model performance for the best performing model-setting pair on edge device.}
\label{tab:qidk_results}
\end{table}
\vspace{-5mm}
Beyond latency, memory constraints further impact edge deployments. As shown in Table \ref{tab: mem_usage}, models consume significantly more memory on servers, with Deepseek-Coder requiring 5385.89MB compared to 1678.18MB on an edge device. Similar trends hold for most other models, achieving around a 5x decrease in memory usage when deployed on the edge.

\noindent
\begin{tcolorbox}[colback=red!5!white,colframe=red!75!black]
\textbf{Main findings of RQ4}: Edge device deployment shows that Deepseek-Coder achieves the best balance between performance and latency and memory usage. While few-shot learning yields better accuracy metrics, it comes with slightly increased latency compared to other settings. We also observe that models consume about 5x less memory on edge devices than on servers due to GGUF.
\end{tcolorbox}

\noindent The replication package containing the code, prompt templates and data is available here \footnote{\url{https://github.com/Raghav010/Small-Models-Big-Tasks}}.

\vspace{-3mm}

\section{Discussion} \label{sec: discussion}

\begin{figure*}
    \centering
    \includegraphics[width=0.8\linewidth]{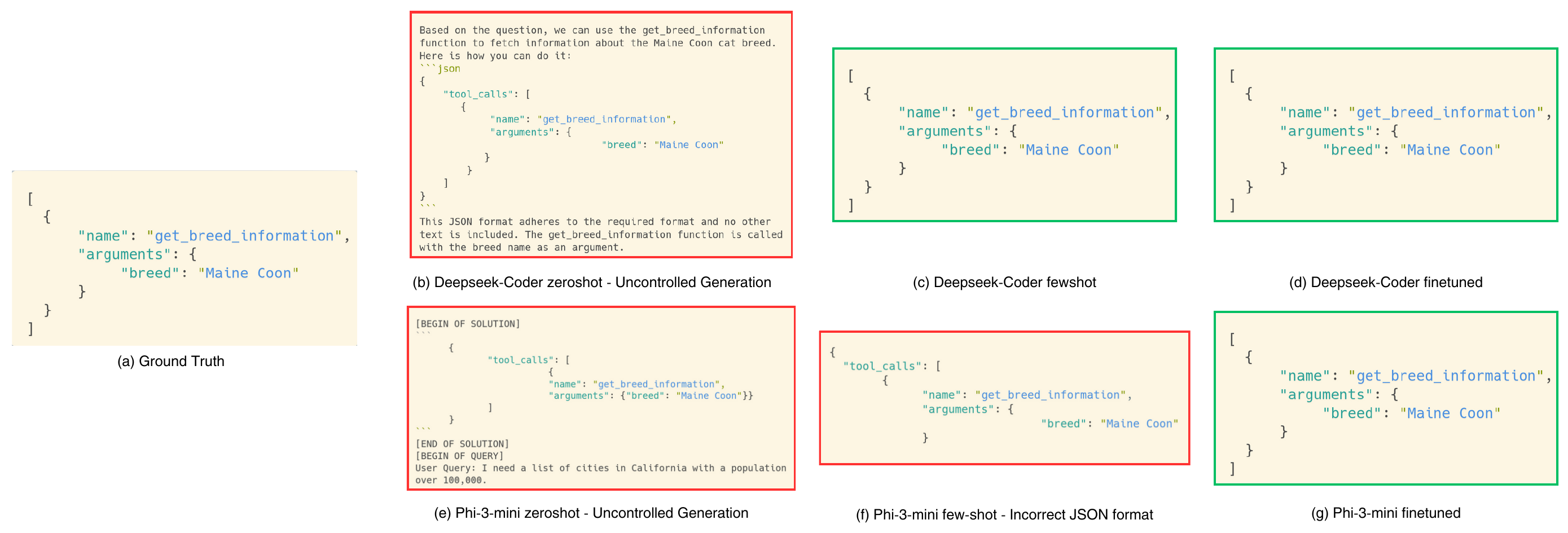}
    \caption{Responses of Deepseek-Coder and Phi-3-mini across settings highlighting incorrect (\textcolor{red}{red}) and correct (\textcolor[HTML]{00BF63}{green}) outputs.}
    \label{fig:examples}
\end{figure*}

\hspace{0.3cm} \textbf{Lessons Learned}: Our study reveals several key insights into the function calling capabilities of SLMs. First, zero-shot and few-shot performance across all evaluated models is generally poor, with Deepseek-Coder being the only model that can generate function calls in a structured format. Finetuning the models significantly improves the model performance. However, we still observe many models with zero values in Table \ref{tab:results}.  Figure \ref{fig:examples} illustrates examples of Deepseek-Coder and Phi-3-mini responses across different settings. Common problems found across different settings include uncontrolled generation and an inability of models to produce responses in the specified format. Figure \ref{fig:examples}(b), \ref{fig:examples}(e) and \ref{fig:examples}(f) illustrate instances where a model generates a correct answer but fails to adhere to the required JSON format or continues generating text beyond the answer.
 
However, these responses do not show any pattern, making it difficult to come up with a systematic method to process these responses for further use. Hence, we do not investigate this further.

Additionally, the vulnerability of models in the few-shot setting to prompt injection remains a concern, as minor perturbations lead to performance degradation. On the other hand fine-tuned models exhibit greater robustness against prompt perturbations, indicating improvements in structured response adherence. This necessitates that more work needs to be done around building small models which can handle these perturbations and also exploring the use of input-validators which can clean the prompt for inaccuracies before passing it on to the model. It would require a mix of SE and NLP expertise to make this possible.

\begin{figure}[H]
    \centering
    \includegraphics[width=0.9\linewidth]{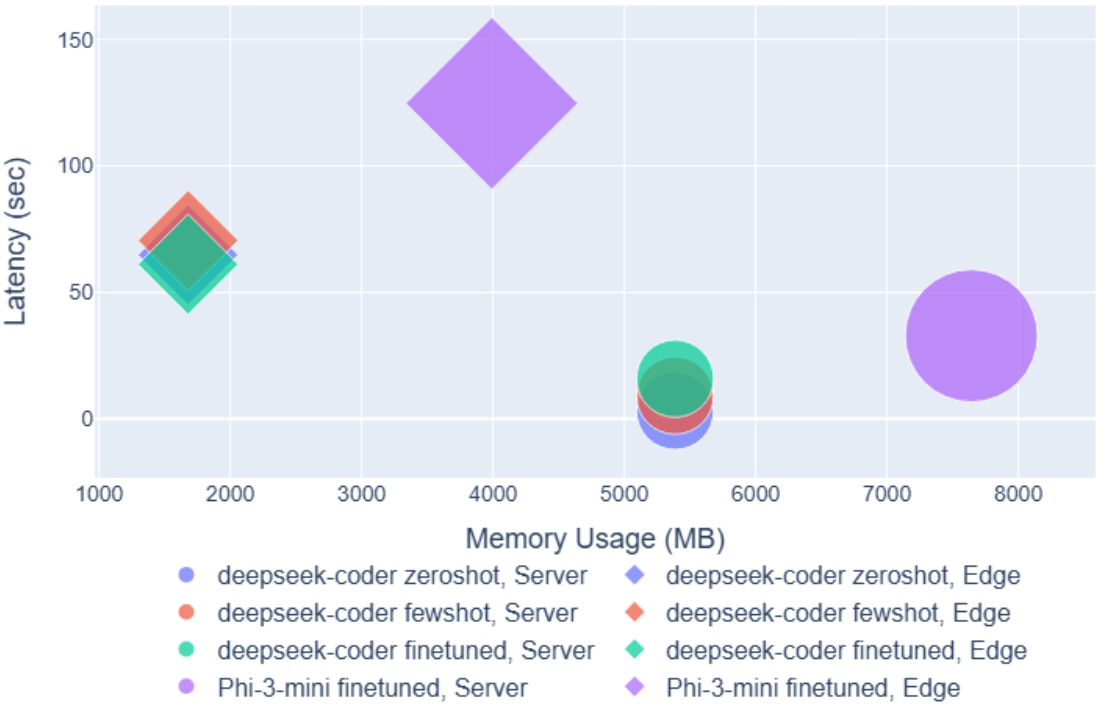}
    \caption{Comparison of latency (in sec) and Memory usage (in MB) of Models on edge and server environments. }
    \label{fig:lat_mem}
\end{figure}

Deploying SLMs on edge devices requires balancing latency, memory, and performance. As shown in Figure \ref{fig:lat_mem}, edge latencies often exceed server latencies by an order of magnitude, necessitating model optimizations and efficient inference techniques like flash attention \cite{dao2022flashattentionfastmemoryefficientexact} or hardware acceleration. Sustained usage also faces power and temperature constraints, which can be mitigated through NPUs and specialized accelerators for improved efficiency and thermal management.

\textbf{Future Research in NLP}: Our findings highlight several avenues for further research in improving SLMs for generation tasks. The issue of uncontrolled generation and improper formatting necessitates advancements in decoding strategies. Research into constrained decoding techniques \cite{wang2024deepedit, hu-etal-2019-improved}, such as grammar-constrained sampling and structured output models, could help mitigate these challenges.


Another point to note is that we experiment with a simple prompt injection attack and observe a significant decline in model performance. A more complex attack such as multi-turn prompt injection \cite{agarwal-etal-2024-prompt} has the potential to manipulate the model into leaking sensitive system information. Hence, it is essential to develop robust defense mechanisms to mitigate the impact of such attacks.

Building task-specific fine-tuned models may present challenges in generalizability across different function calling domains. Future research could explore cross-task applicability, and leveraging synthetic data generation for domain specific datasets.

\begin{table}[H]
\resizebox{\linewidth}{!}{
\begin{tabular}{ccc}
\hline
\textbf{Model} & \textbf{Server (fp16)} & \textbf{Edge (GGUF)} \\ \hline
Deepseek-coder-1.3B-instruct & 2,570.24 & 1,678.18 \\
Phi-3-mini-4k-instruct & 7,642.16 & 3,990.97 \\
Phi-2 & 5,304.32 & 2,443.01 \\
Starcoder2-3B & 5,775.36 & 2,051.05 \\
Stable-code-3B & 5,335.04 & 3,058.95 \\ \hline
\end{tabular}
}
\caption{Memory usage (MB) for models across settings.}
\label{tab: mem_usage}
\end{table}

\textbf{Future Research for SE}:
SLMs present promising opportunities for efficient and scalable solutions in software engineering. Finetuned models have demonstrated strong performance in function calling tasks, highlighting their potential for real-world applications. Additionally, our study shows prompt injection has minimal impact on their effectiveness. Researchers should focus on developing more robust fine-tuned SLMs that can handle diverse scenarios, ensuring greater reliability and adaptability in function call generation. Further studies on generalizability across different SE lifecycle tasks and resilience against various attacks can enhance the robustness of these models, making them more viable for real-world deployment. This research will help bridge the gap between experimentation and practical applications.

SLMs also present an opportunity is in terms of re-imagining the integration of models with software systems to make them more efficient. 

Most LLM-powered applications rely on a single LLM to handle and solve multitude of tasks. In contrast, SLM research has focused on building task-specific models that execute specialized functions with greater efficiency. This can be leveraged to engineer sustainable software systems by decomposing large complex tasks and routing them to various specialized SLMs, maintaining overall performance while significantly reducing computational costs.


\begin{table}[H]
\resizebox{0.8\linewidth}{!}{
\begin{tabular}{ccccc}
\hline
\textbf{Setting} & \textbf{Model} & \textbf{Zero-shot} & \textbf{Few-shot} & \textbf{Finetuned} \\ \hline
\multirow{5}{*}{\begin{tabular}[c]{@{}c@{}}Server\\ (fp16)\end{tabular}} & Deepseek-coder-1.3B-instruct & 5.59 & 3.22 & 3.91 \\
 & Phi-3-mini-4k-instruct & 35.23 & 39.14 & 32.88 \\
 & Phi-2 & 22.19 & 28.97 & 62.40 \\
 & Starcoder2-3B & 32.05 & 33.25 & 60.32 \\
 & Stable-code-3B & 22.95 & 27.67 & 31.91 \\ \hline
\multirow{5}{*}{\begin{tabular}[c]{@{}c@{}}Edge\\ (GGUF)\end{tabular}} & Deepseek-coder-1.3B-instruct & 64.87 & 70.55 & 61.10 \\
 & Phi-3-mini-4k-instruct & 364.51 & 335.58 & 124.76 \\
 & Phi-2 & 153.67 & 250.77 & 140.29 \\
 & Starcoder2-3B & 246.63 & 262.92 & 267.65 \\
 & Stable-code-3B & 214.89 & 265.90 & 175.77 \\ \hline
\end{tabular}
}
 \caption{Latency of models (sec) for models across settings.}
 \label{tab: latency}
\end{table}
\vspace{-3mm}

This shift towards task-specific SLMs not only enhances efficiency but also opens avenues for hybrid deployment strategies\cite{Hao2024HybridSA}. Techniques such as edge-cloud collaboration, where initial inference occurs on-device and complex queries are offloaded to the cloud, can reduce latency while maintaining robustness. Techniques such as model compression \cite{zhu2024survey}, quantization \cite{dettmers2022gpt3}, and knowledge distillation \cite{gu2023knowledge} can help optimize computational efficiency, enable faster inference and reduce reliance on cloud resources. However, their effectiveness for software engineering tasks is underexplored, presenting a direction for future research.

\textbf{Implications for Practice}: Our findings indicate that deploying SLMs for function calling in real-world applications requires careful consideration of performance, reliability, and integration challenges.  Real-world environments impose higher expectations on performance and reliability, as application developers require models that can be safely integrated into production systems. 
Nonetheless, SLMs offer distinct advantages, particularly in real-time applications where latency is crucial. This not only reduces inference costs but also supports Greening AI initiatives by lowering energy consumption and carbon footprint compared to large cloud-based models.
However, one challenge remains-inconsistent adherence to structured formats, which presents challenges for integration into production systems. 
Most models do not adhere to the specified format, necessitating engineering solutions for enhancing adaptability. Approaches such as wrappers that enforce structured outputs, response format checkers, and guardrails can help mitigate malformed outputs.
Prompt engineering, which involves carefully designing instructions to align with model-specific prompting formats, can improve structured output generation. 
However, perturbations to the prompt via prompt injection attacks can render the outputs unusable.
In contrast, tools such as Typechat \footnote{https://microsoft.github.io/TypeChat/} and Guidance \footnote{https://github.com/guidance-ai/guidance} can be used to steer the generative models to generate structured outputs. TypeChat enables TypeScript-based interactions with models, using re-prompting and the TypeScript compiler for output validation. In contrast, Guidance combines prompt engineering, constrained token generation, and re-prompting to improve control and precision in model responses. We believe that developers can greatly benefit from these tools for generating better responses and refining the malformed ones.

A key deployment consideration is choosing between few-shot prompting and fine-tuning. Few-shot prompting offers flexibility, allowing easy adaptation to backend updates through prompt modifications. Fine-tuning enhances task performance and robustness to prompt perturbations but demands higher compute and maintenance. The choice depends on balancing adaptability with performance needs.


Despite the potential of SLMs in function-calling systems, their adoption remains limited due to several challenges like lack of efficient deployment frameworks, and the heterogeneous nature of edge computing devices, which complicate the integration of these models into real-world systems \cite{zheng2024reviewedgelargelanguage}. Additionally, ensuring reliable performance and low-latency execution in dynamic, resource-constrained environments is a persistent hurdle. Tackling these challenges calls for a stronger collaboration between the SE and NLP communities.

\vspace{-2mm}
\section{Threats to validity} \label{sec: threats}

In this section we discuss the threats to validity of our study. We follow the categorization provided by Wohlin et al. \cite{wohlin2012experimentation} and explain how we try to mitigate them.

\textbf{Internal Validity}:  The selection of models from the EvalPlus leaderboard was based on their demonstrated coding abilities, reducing the risk of bias in model selection. This ensures that our analysis is focused on state-of-the-art models. A threat to validity can arise from the metrics used in the study. To the best of our knowledge, no standard metrics exist to extensively evaluate function calls. Existing metrics like ROUGE, BLEU, and METEOR focus on the textual syntax and semantics, making them unfit for our usecase. Hence, we define metrics that evaluate various aspects of a function definition.

\textbf{External Validity}: The rapid release of newer SLMs with advanced capabilities make it unfeasible to consider all the models for the study. Hence, we carefully select the top performing models from the EvalPlus leaderboard which match our requirements. Another threat to validity is the possibility that the models may have been exposed to the test data during training. However, since most models do not publicly disclose detailed information about their training data, we are unable to mitigate this data leaking issue. We exclusively consider the QIDK device for our edge experiments. However, performance results may be different between other edge devices with different hardware configurations. Future work could explore how these models perform on a broader range of edge devices.

 \textbf{Construct Validity}: A potential threat to construct validity can arise from the dataset used to evaluate the models. To mitigate this, we select the salesforce-xlam function calling dataset which contains a diverse set of real-world function calls from domains such as finance, technology, health care, sports and more. We choose 100 datapoints from the dataset for our edge device experiments. A threat to validity might stem from this if the subset is not a representative of the whole dataset. To mitigate this, we randomly select the datapoints to ensure a balanced sample.
Additionally, the metrics used in the study are designed to capture various aspects of function call generation, providing a comprehensive evaluation of the models’ capabilities across multiple dimensions. 
\vspace{-3.5mm}
\section{Conclusion} \label{sec: conclusion}
In this study, we explore the ability of Small Language Models to generate function calls. We evaluate five top ranked SLMs from the EvalPlus leaderboard across various experimental settings such as zero-shot, few-shot, finetuning, and on an edge device. In addition, we assess the robustness of these models by performing prompt injection experiments. We find that SLMs struggle to generate function calls autonomously, their performance improves with post-processing, albeit with some degradation due to prompt injection. Furthermore, we observe that high performance of a model on a server does not necessarily translate into edge device settings. Furthermore, we provide the finetuned models from our study and outline a path forward for the NLP and SE communities to enhance function call generation in practical scenarios.

Future work can involve extending this study to language specific function call generation, providing insights on model performance across programming languages. 
Exploring additional edge devices and assessing various adversarial attacks would provide a more comprehensive robustness evaluation.
Another potential direction is leveraging LLM-based agents and multi-agent systems to generate function calls and execute them. 
\vspace{-3mm}

\section*{Acknowledgements}
We sincerely thank Qualcomm Inc. and AlphaGrep for their generous funding and support for this study through supporting IIIT-Hyderabad, India. 
We also thank Sujay Belsare and Ananth Yegavakota for their help with edge device experiments. Additionally, we thank Kunal Bhosikar, Hiya Bhatt, Shrikara A, and other members of SERC and Precog groups, IIIT-Hyderabad for their valuable feedback.

\bibliographystyle{ACM-Reference-Format}
\bibliography{main}


\begin{thebibliography}{62}


\ifx \showCODEN    \undefined \def \showCODEN     #1{\unskip}     \fi
\ifx \showISBNx    \undefined \def \showISBNx     #1{\unskip}     \fi
\ifx \showISBNxiii \undefined \def \showISBNxiii  #1{\unskip}     \fi
\ifx \showISSN     \undefined \def \showISSN      #1{\unskip}     \fi
\ifx \showLCCN     \undefined \def \showLCCN      #1{\unskip}     \fi
\ifx \shownote     \undefined \def \shownote      #1{#1}          \fi
\ifx \showarticletitle \undefined \def \showarticletitle #1{#1}   \fi
\ifx \showURL      \undefined \def \showURL       {\relax}        \fi
\providecommand\bibfield[2]{#2}
\providecommand\bibinfo[2]{#2}
\providecommand\natexlab[1]{#1}
\providecommand\showeprint[2][]{arXiv:#2}

\bibitem[Abdelaziz et~al\mbox{.}(2024)]%
        {abdelaziz2024granite}
\bibfield{author}{\bibinfo{person}{Ibrahim Abdelaziz}, \bibinfo{person}{Kinjal Basu}, \bibinfo{person}{Mayank Agarwal}, \bibinfo{person}{Sadhana Kumaravel}, \bibinfo{person}{Matthew Stallone}, \bibinfo{person}{Rameswar Panda}, \bibinfo{person}{Yara Rizk}, \bibinfo{person}{GP Bhargav}, \bibinfo{person}{Maxwell Crouse}, \bibinfo{person}{Chulaka Gunasekara}, {et~al\mbox{.}}} \bibinfo{year}{2024}\natexlab{}.
\newblock \showarticletitle{Granite-function calling model: Introducing function calling abilities via multi-task learning of granular tasks}.
\newblock \bibinfo{journal}{\emph{arXiv preprint arXiv:2407.00121}} (\bibinfo{year}{2024}).
\newblock


\bibitem[Abdin et~al\mbox{.}(2024)]%
        {abdin2024phi3technicalreporthighly}
\bibfield{author}{\bibinfo{person}{Marah Abdin}, \bibinfo{person}{Jyoti Aneja}, \bibinfo{person}{Hany Awadalla}, \bibinfo{person}{Ahmed Awadallah}, \bibinfo{person}{Ammar~Ahmad Awan}, \bibinfo{person}{Nguyen Bach}, \bibinfo{person}{Amit Bahree}, {and} \bibinfo{person}{Arash~Bakhtiari et al.}} \bibinfo{year}{2024}\natexlab{}.
\newblock \bibinfo{title}{Phi-3 Technical Report: A Highly Capable Language Model Locally on Your Phone}.
\newblock
\showeprint[arxiv]{2404.14219}~[cs.CL]
\urldef\tempurl%
\url{https://arxiv.org/abs/2404.14219}
\showURL{%
\tempurl}


\bibitem[Agarwal et~al\mbox{.}(2024)]%
        {agarwal-etal-2024-prompt}
\bibfield{author}{\bibinfo{person}{Divyansh Agarwal}, \bibinfo{person}{Alexander Fabbri}, \bibinfo{person}{Ben Risher}, \bibinfo{person}{Philippe Laban}, \bibinfo{person}{Shafiq Joty}, {and} \bibinfo{person}{Chien-Sheng Wu}.} \bibinfo{year}{2024}\natexlab{}.
\newblock \showarticletitle{Prompt Leakage effect and mitigation strategies for multi-turn {LLM} Applications}. In \bibinfo{booktitle}{\emph{Proceedings of the 2024 Conference on Empirical Methods in Natural Language Processing: Industry Track}}, \bibfield{editor}{\bibinfo{person}{Franck Dernoncourt}, \bibinfo{person}{Daniel Preo{\c{t}}iuc-Pietro}, {and} \bibinfo{person}{Anastasia Shimorina}} (Eds.). \bibinfo{publisher}{Association for Computational Linguistics}, \bibinfo{address}{Miami, Florida, US}, \bibinfo{pages}{1255--1275}.
\newblock
\href{https://doi.org/10.18653/v1/2024.emnlp-industry.94}{doi:\nolinkurl{10.18653/v1/2024.emnlp-industry.94}}


\bibitem[Ahmad et~al\mbox{.}(2021)]%
        {ahmad2021unified}
\bibfield{author}{\bibinfo{person}{Wasi~Uddin Ahmad}, \bibinfo{person}{Saikat Chakraborty}, \bibinfo{person}{Baishakhi Ray}, {and} \bibinfo{person}{Kai-Wei Chang}.} \bibinfo{year}{2021}\natexlab{}.
\newblock \showarticletitle{Unified pre-training for program understanding and generation}.
\newblock \bibinfo{journal}{\emph{arXiv preprint arXiv:2103.06333}} (\bibinfo{year}{2021}).
\newblock


\bibitem[Akbar et~al\mbox{.}(2024)]%
        {akbar2024devops}
\bibfield{author}{\bibinfo{person}{Muhammad~Azeem Akbar}, \bibinfo{person}{Arif~Ali Khan}, \bibinfo{person}{Najmul Islam}, {and} \bibinfo{person}{Sajjad Mahmood}.} \bibinfo{year}{2024}\natexlab{}.
\newblock \showarticletitle{DevOps project management success factors: A decision-making framework}.
\newblock \bibinfo{journal}{\emph{Software: Practice and Experience}} \bibinfo{volume}{54}, \bibinfo{number}{2} (\bibinfo{year}{2024}), \bibinfo{pages}{257--280}.
\newblock


\bibitem[Banerjee and Lavie(2004)]%
        {banerjee2004meteor}
\bibfield{author}{\bibinfo{person}{Satanjeev Banerjee} {and} \bibinfo{person}{Alon Lavie}.} \bibinfo{year}{2004}\natexlab{}.
\newblock \showarticletitle{Meteor: an automatic metric for MT evaluation with high levels of correlation with human judgments}.
\newblock \bibinfo{journal}{\emph{Proceedings of ACL-WMT}} (\bibinfo{year}{2004}), \bibinfo{pages}{65--72}.
\newblock


\bibitem[Basu et~al\mbox{.}(2024)]%
        {basu2024api}
\bibfield{author}{\bibinfo{person}{Kinjal Basu}, \bibinfo{person}{Ibrahim Abdelaziz}, \bibinfo{person}{Subhajit Chaudhury}, \bibinfo{person}{Soham Dan}, \bibinfo{person}{Maxwell Crouse}, \bibinfo{person}{Asim Munawar}, \bibinfo{person}{Sadhana Kumaravel}, \bibinfo{person}{Vinod Muthusamy}, \bibinfo{person}{Pavan Kapanipathi}, {and} \bibinfo{person}{Luis~A Lastras}.} \bibinfo{year}{2024}\natexlab{}.
\newblock \showarticletitle{API-BLEND: A Comprehensive Corpora for Training and Benchmarking API LLMs}.
\newblock \bibinfo{journal}{\emph{arXiv preprint arXiv:2402.15491}} (\bibinfo{year}{2024}).
\newblock


\bibitem[Branch et~al\mbox{.}(2022)]%
        {branch2022evaluating}
\bibfield{author}{\bibinfo{person}{Hezekiah~J Branch}, \bibinfo{person}{Jonathan~Rodriguez Cefalu}, \bibinfo{person}{Jeremy McHugh}, \bibinfo{person}{Leyla Hujer}, \bibinfo{person}{Aditya Bahl}, \bibinfo{person}{Daniel del~Castillo Iglesias}, \bibinfo{person}{Ron Heichman}, {and} \bibinfo{person}{Ramesh Darwishi}.} \bibinfo{year}{2022}\natexlab{}.
\newblock \showarticletitle{Evaluating the susceptibility of pre-trained language models via handcrafted adversarial examples}.
\newblock \bibinfo{journal}{\emph{arXiv preprint arXiv:2209.02128}} (\bibinfo{year}{2022}).
\newblock


\bibitem[Brown et~al\mbox{.}(2020)]%
        {10.5555/3495724.3495883}
\bibfield{author}{\bibinfo{person}{Tom~B. Brown}, \bibinfo{person}{Benjamin Mann}, \bibinfo{person}{Nick Ryder}, \bibinfo{person}{Melanie Subbiah}, {and} \bibinfo{person}{et al.}} \bibinfo{year}{2020}\natexlab{}.
\newblock \showarticletitle{Language models are few-shot learners}. In \bibinfo{booktitle}{\emph{Proceedings of the 34th International Conference on Neural Information Processing Systems}} (Vancouver, BC, Canada) \emph{(\bibinfo{series}{NIPS '20})}. \bibinfo{publisher}{Curran Associates Inc.}, \bibinfo{address}{Red Hook, NY, USA}, Article \bibinfo{articleno}{159}, \bibinfo{numpages}{25}~pages.
\newblock
\showISBNx{9781713829546}


\bibitem[Busari and Letier(2017)]%
        {busari2017radar}
\bibfield{author}{\bibinfo{person}{Saheed~A Busari} {and} \bibinfo{person}{Emmanuel Letier}.} \bibinfo{year}{2017}\natexlab{}.
\newblock \showarticletitle{Radar: A lightweight tool for requirements and architecture decision analysis}. In \bibinfo{booktitle}{\emph{2017 IEEE/ACM 39th International Conference on Software Engineering (ICSE)}}. IEEE, \bibinfo{pages}{552--562}.
\newblock


\bibitem[Caldiera and Rombach(1994)]%
        {caldiera1994goal}
\bibfield{author}{\bibinfo{person}{Victor R Basili1~Gianluigi Caldiera} {and} \bibinfo{person}{H~Dieter Rombach}.} \bibinfo{year}{1994}\natexlab{}.
\newblock \showarticletitle{The goal question metric approach}.
\newblock \bibinfo{journal}{\emph{Encyclopedia of software engineering}} (\bibinfo{year}{1994}), \bibinfo{pages}{528--532}.
\newblock


\bibitem[Chen et~al\mbox{.}(2025)]%
        {chen2025empiricalstudychallengesllm}
\bibfield{author}{\bibinfo{person}{Xiang Chen}, \bibinfo{person}{Chaoyang Gao}, \bibinfo{person}{Chunyang Chen}, \bibinfo{person}{Guangbei Zhang}, {and} \bibinfo{person}{Yong Liu}.} \bibinfo{year}{2025}\natexlab{}.
\newblock \bibinfo{title}{An Empirical Study on Challenges for LLM Application Developers}.
\newblock
\showeprint[arxiv]{2408.05002}~[cs.SE]
\urldef\tempurl%
\url{https://arxiv.org/abs/2408.05002}
\showURL{%
\tempurl}


\bibitem[Dao et~al\mbox{.}(2022)]%
        {dao2022flashattentionfastmemoryefficientexact}
\bibfield{author}{\bibinfo{person}{Tri Dao}, \bibinfo{person}{Daniel~Y. Fu}, \bibinfo{person}{Stefano Ermon}, \bibinfo{person}{Atri Rudra}, {and} \bibinfo{person}{Christopher Ré}.} \bibinfo{year}{2022}\natexlab{}.
\newblock \bibinfo{title}{FlashAttention: Fast and Memory-Efficient Exact Attention with IO-Awareness}.
\newblock
\showeprint[arxiv]{2205.14135}~[cs.LG]
\urldef\tempurl%
\url{https://arxiv.org/abs/2205.14135}
\showURL{%
\tempurl}


\bibitem[Dettmers et~al\mbox{.}(2022)]%
        {dettmers2022gpt3}
\bibfield{author}{\bibinfo{person}{Tim Dettmers}, \bibinfo{person}{Mike Lewis}, \bibinfo{person}{Younes Belkada}, {and} \bibinfo{person}{Luke Zettlemoyer}.} \bibinfo{year}{2022}\natexlab{}.
\newblock \showarticletitle{Gpt3. int8 (): 8-bit matrix multiplication for transformers at scale}.
\newblock \bibinfo{journal}{\emph{Advances in Neural Information Processing Systems}}  \bibinfo{volume}{35} (\bibinfo{year}{2022}), \bibinfo{pages}{30318--30332}.
\newblock


\bibitem[Gerganov({[n.\,d.]})]%
        {ggml}
\bibfield{author}{\bibinfo{person}{Georgi Gerganov}.} \bibinfo{year}{[n.\,d.]}\natexlab{}.
\newblock \bibinfo{title}{GGML}.
\newblock
\urldef\tempurl%
\url{https://github.com/ggerganov/ggml}
\showURL{%
\tempurl}


\bibitem[Gerganov(2023)]%
        {gguf}
\bibfield{author}{\bibinfo{person}{Georgi Gerganov}.} \bibinfo{year}{2023}\natexlab{}.
\newblock \bibinfo{title}{GGUF}.
\newblock
\urldef\tempurl%
\url{https://github.com/ggerganov/ggml/blob/master/docs/gguf.md}
\showURL{%
\tempurl}


\bibitem[Greshake et~al\mbox{.}(2023)]%
        {greshake2023not}
\bibfield{author}{\bibinfo{person}{Kai Greshake}, \bibinfo{person}{Sahar Abdelnabi}, \bibinfo{person}{Shailesh Mishra}, \bibinfo{person}{Christoph Endres}, \bibinfo{person}{Thorsten Holz}, {and} \bibinfo{person}{Mario Fritz}.} \bibinfo{year}{2023}\natexlab{}.
\newblock \showarticletitle{Not what you've signed up for: Compromising real-world llm-integrated applications with indirect prompt injection}. In \bibinfo{booktitle}{\emph{Proceedings of the 16th ACM Workshop on Artificial Intelligence and Security}}. \bibinfo{pages}{79--90}.
\newblock


\bibitem[Gu et~al\mbox{.}(2023)]%
        {gu2023knowledge}
\bibfield{author}{\bibinfo{person}{Yuxian Gu}, \bibinfo{person}{Li Dong}, \bibinfo{person}{Furu Wei}, {and} \bibinfo{person}{Minlie Huang}.} \bibinfo{year}{2023}\natexlab{}.
\newblock \showarticletitle{Knowledge distillation of large language models}.
\newblock \bibinfo{journal}{\emph{arXiv preprint arXiv:2306.08543}} (\bibinfo{year}{2023}).
\newblock


\bibitem[Guo et~al\mbox{.}(2024b)]%
        {guo2024deepseekcoderlargelanguagemodel}
\bibfield{author}{\bibinfo{person}{Daya Guo}, \bibinfo{person}{Qihao Zhu}, \bibinfo{person}{Dejian Yang}, \bibinfo{person}{Zhenda Xie}, \bibinfo{person}{Kai Dong}, \bibinfo{person}{Wentao Zhang}, \bibinfo{person}{Guanting Chen}, \bibinfo{person}{Xiao Bi}, \bibinfo{person}{Y. Wu}, \bibinfo{person}{Y.~K. Li}, \bibinfo{person}{Fuli Luo}, \bibinfo{person}{Yingfei Xiong}, {and} \bibinfo{person}{Wenfeng Liang}.} \bibinfo{year}{2024}\natexlab{b}.
\newblock \bibinfo{title}{DeepSeek-Coder: When the Large Language Model Meets Programming -- The Rise of Code Intelligence}.
\newblock
\showeprint[arxiv]{2401.14196}~[cs.SE]
\urldef\tempurl%
\url{https://arxiv.org/abs/2401.14196}
\showURL{%
\tempurl}


\bibitem[Guo et~al\mbox{.}(2024a)]%
        {guo2024stabletoolbench}
\bibfield{author}{\bibinfo{person}{Zhicheng Guo}, \bibinfo{person}{Sijie Cheng}, \bibinfo{person}{Hao Wang}, \bibinfo{person}{Shihao Liang}, \bibinfo{person}{Yujia Qin}, \bibinfo{person}{Peng Li}, \bibinfo{person}{Zhiyuan Liu}, \bibinfo{person}{Maosong Sun}, {and} \bibinfo{person}{Yang Liu}.} \bibinfo{year}{2024}\natexlab{a}.
\newblock \showarticletitle{StableToolBench: Towards Stable Large-Scale Benchmarking on Tool Learning of Large Language Models}.
\newblock \bibinfo{journal}{\emph{arXiv preprint arXiv:2403.07714}} (\bibinfo{year}{2024}).
\newblock


\bibitem[Hao et~al\mbox{.}(2024)]%
        {Hao2024HybridSA}
\bibfield{author}{\bibinfo{person}{Zixu Hao}, \bibinfo{person}{Huiqiang Jiang}, \bibinfo{person}{Shiqi Jiang}, \bibinfo{person}{Ju Ren}, {and} \bibinfo{person}{Ting Cao}.} \bibinfo{year}{2024}\natexlab{}.
\newblock \showarticletitle{Hybrid SLM and LLM for Edge-Cloud Collaborative Inference}.
\newblock \bibinfo{journal}{\emph{Proceedings of the Workshop on Edge and Mobile Foundation Models}} (\bibinfo{year}{2024}).
\newblock
\urldef\tempurl%
\url{https://api.semanticscholar.org/CorpusID:270405086}
\showURL{%
\tempurl}


\bibitem[Hu et~al\mbox{.}(2021)]%
        {hu2021lora}
\bibfield{author}{\bibinfo{person}{Edward~J Hu}, \bibinfo{person}{Yelong Shen}, \bibinfo{person}{Phillip Wallis}, \bibinfo{person}{Zeyuan Allen-Zhu}, \bibinfo{person}{Yuanzhi Li}, \bibinfo{person}{Shean Wang}, \bibinfo{person}{Lu Wang}, {and} \bibinfo{person}{Weizhu Chen}.} \bibinfo{year}{2021}\natexlab{}.
\newblock \showarticletitle{Lora: Low-rank adaptation of large language models}.
\newblock \bibinfo{journal}{\emph{arXiv preprint arXiv:2106.09685}} (\bibinfo{year}{2021}).
\newblock


\bibitem[Hu et~al\mbox{.}(2019)]%
        {hu-etal-2019-improved}
\bibfield{author}{\bibinfo{person}{J.~Edward Hu}, \bibinfo{person}{Huda Khayrallah}, \bibinfo{person}{Ryan Culkin}, \bibinfo{person}{Patrick Xia}, \bibinfo{person}{Tongfei Chen}, \bibinfo{person}{Matt Post}, {and} \bibinfo{person}{Benjamin Van~Durme}.} \bibinfo{year}{2019}\natexlab{}.
\newblock \showarticletitle{Improved Lexically Constrained Decoding for Translation and Monolingual Rewriting}. In \bibinfo{booktitle}{\emph{Proceedings of the 2019 Conference of the North {A}merican Chapter of the Association for Computational Linguistics: Human Language Technologies, Volume 1 (Long and Short Papers)}}, \bibfield{editor}{\bibinfo{person}{Jill Burstein}, \bibinfo{person}{Christy Doran}, {and} \bibinfo{person}{Thamar Solorio}} (Eds.). \bibinfo{publisher}{Association for Computational Linguistics}, \bibinfo{address}{Minneapolis, Minnesota}, \bibinfo{pages}{839--850}.
\newblock
\href{https://doi.org/10.18653/v1/N19-1090}{doi:\nolinkurl{10.18653/v1/N19-1090}}


\bibitem[Javaheripi et~al\mbox{.}(2023)]%
        {javaheripi2023phi}
\bibfield{author}{\bibinfo{person}{Mojan Javaheripi}, \bibinfo{person}{S{\'e}bastien Bubeck}, \bibinfo{person}{Marah Abdin}, \bibinfo{person}{Jyoti Aneja}, \bibinfo{person}{Sebastien Bubeck}, \bibinfo{person}{Caio C{\'e}sar~Teodoro Mendes}, \bibinfo{person}{Weizhu Chen}, \bibinfo{person}{Allie Del~Giorno}, \bibinfo{person}{Ronen Eldan}, \bibinfo{person}{Sivakanth Gopi}, {et~al\mbox{.}}} \bibinfo{year}{2023}\natexlab{}.
\newblock \showarticletitle{Phi-2: The surprising power of small language models}.
\newblock \bibinfo{journal}{\emph{Microsoft Research Blog}} \bibinfo{volume}{1}, \bibinfo{number}{3} (\bibinfo{year}{2023}), \bibinfo{pages}{3}.
\newblock


\bibitem[Jiang et~al\mbox{.}(2021)]%
        {jiang2021cure}
\bibfield{author}{\bibinfo{person}{Nan Jiang}, \bibinfo{person}{Thibaud Lutellier}, {and} \bibinfo{person}{Lin Tan}.} \bibinfo{year}{2021}\natexlab{}.
\newblock \showarticletitle{Cure: Code-aware neural machine translation for automatic program repair}. In \bibinfo{booktitle}{\emph{2021 IEEE/ACM 43rd International Conference on Software Engineering (ICSE)}}. IEEE, \bibinfo{pages}{1161--1173}.
\newblock


\bibitem[Kojima et~al\mbox{.}(2022)]%
        {kojima2022large}
\bibfield{author}{\bibinfo{person}{Takeshi Kojima}, \bibinfo{person}{Shixiang~Shane Gu}, \bibinfo{person}{Machel Reid}, \bibinfo{person}{Yutaka Matsuo}, {and} \bibinfo{person}{Yusuke Iwasawa}.} \bibinfo{year}{2022}\natexlab{}.
\newblock \showarticletitle{Large language models are zero-shot reasoners}.
\newblock \bibinfo{journal}{\emph{Advances in neural information processing systems}}  \bibinfo{volume}{35} (\bibinfo{year}{2022}), \bibinfo{pages}{22199--22213}.
\newblock


\bibitem[Lester et~al\mbox{.}(2021)]%
        {lester-etal-2021-power}
\bibfield{author}{\bibinfo{person}{Brian Lester}, \bibinfo{person}{Rami Al-Rfou}, {and} \bibinfo{person}{Noah Constant}.} \bibinfo{year}{2021}\natexlab{}.
\newblock \showarticletitle{The Power of Scale for Parameter-Efficient Prompt Tuning}. In \bibinfo{booktitle}{\emph{Proceedings of the 2021 Conference on Empirical Methods in Natural Language Processing}}, \bibfield{editor}{\bibinfo{person}{Marie-Francine Moens}, \bibinfo{person}{Xuanjing Huang}, \bibinfo{person}{Lucia Specia}, {and} \bibinfo{person}{Scott Wen-tau Yih}} (Eds.). \bibinfo{publisher}{Association for Computational Linguistics}, \bibinfo{address}{Online and Punta Cana, Dominican Republic}, \bibinfo{pages}{3045--3059}.
\newblock
\href{https://doi.org/10.18653/v1/2021.emnlp-main.243}{doi:\nolinkurl{10.18653/v1/2021.emnlp-main.243}}


\bibitem[Li and Liang(2021)]%
        {li-liang-2021-prefix}
\bibfield{author}{\bibinfo{person}{Xiang~Lisa Li} {and} \bibinfo{person}{Percy Liang}.} \bibinfo{year}{2021}\natexlab{}.
\newblock \showarticletitle{Prefix-Tuning: Optimizing Continuous Prompts for Generation}. In \bibinfo{booktitle}{\emph{Proceedings of the 59th Annual Meeting of the Association for Computational Linguistics and the 11th International Joint Conference on Natural Language Processing (Volume 1: Long Papers)}}, \bibfield{editor}{\bibinfo{person}{Chengqing Zong}, \bibinfo{person}{Fei Xia}, \bibinfo{person}{Wenjie Li}, {and} \bibinfo{person}{Roberto Navigli}} (Eds.). \bibinfo{publisher}{Association for Computational Linguistics}, \bibinfo{address}{Online}, \bibinfo{pages}{4582--4597}.
\newblock
\href{https://doi.org/10.18653/v1/2021.acl-long.353}{doi:\nolinkurl{10.18653/v1/2021.acl-long.353}}


\bibitem[Li et~al\mbox{.}(2023)]%
        {li2023cctest}
\bibfield{author}{\bibinfo{person}{Zongjie Li}, \bibinfo{person}{Chaozheng Wang}, \bibinfo{person}{Zhibo Liu}, \bibinfo{person}{Haoxuan Wang}, \bibinfo{person}{Dong Chen}, \bibinfo{person}{Shuai Wang}, {and} \bibinfo{person}{Cuiyun Gao}.} \bibinfo{year}{2023}\natexlab{}.
\newblock \showarticletitle{Cctest: Testing and repairing code completion systems}. In \bibinfo{booktitle}{\emph{2023 IEEE/ACM 45th International Conference on Software Engineering (ICSE)}}. IEEE, \bibinfo{pages}{1238--1250}.
\newblock


\bibitem[Lin(2004)]%
        {lin2004rouge}
\bibfield{author}{\bibinfo{person}{Chin-Yew Lin}.} \bibinfo{year}{2004}\natexlab{}.
\newblock \showarticletitle{Rouge: A package for automatic evaluation of summaries}. In \bibinfo{booktitle}{\emph{Text summarization branches out}}. \bibinfo{pages}{74--81}.
\newblock


\bibitem[Lin et~al\mbox{.}(2015)]%
        {lin2015multi}
\bibfield{author}{\bibinfo{person}{Yishuai Lin}, \bibinfo{person}{Philippe Descamps}, \bibinfo{person}{Nicolas Gaud}, \bibinfo{person}{Vincent Hilaire}, {and} \bibinfo{person}{Abderrafiaa Koukam}.} \bibinfo{year}{2015}\natexlab{}.
\newblock \showarticletitle{Multi-agent system for intelligent scrum project management}.
\newblock \bibinfo{journal}{\emph{Integrated Computer-Aided Engineering}} \bibinfo{volume}{22}, \bibinfo{number}{3} (\bibinfo{year}{2015}), \bibinfo{pages}{281--296}.
\newblock


\bibitem[Liu et~al\mbox{.}(2023)]%
        {evalplus}
\bibfield{author}{\bibinfo{person}{Jiawei Liu}, \bibinfo{person}{Chunqiu~Steven Xia}, \bibinfo{person}{Yuyao Wang}, {and} \bibinfo{person}{Lingming Zhang}.} \bibinfo{year}{2023}\natexlab{}.
\newblock \showarticletitle{Is Your Code Generated by Chat{GPT} Really Correct? Rigorous Evaluation of Large Language Models for Code Generation}. In \bibinfo{booktitle}{\emph{Thirty-seventh Conference on Neural Information Processing Systems}}.
\newblock
\urldef\tempurl%
\url{https://openreview.net/forum?id=1qvx610Cu7}
\showURL{%
\tempurl}


\bibitem[Liu et~al\mbox{.}(2024b)]%
        {liu2024formalizing}
\bibfield{author}{\bibinfo{person}{Yupei Liu}, \bibinfo{person}{Yuqi Jia}, \bibinfo{person}{Runpeng Geng}, \bibinfo{person}{Jinyuan Jia}, {and} \bibinfo{person}{Neil~Zhenqiang Gong}.} \bibinfo{year}{2024}\natexlab{b}.
\newblock \showarticletitle{Formalizing and benchmarking prompt injection attacks and defenses}. In \bibinfo{booktitle}{\emph{33rd USENIX Security Symposium (USENIX Security 24)}}. \bibinfo{pages}{1831--1847}.
\newblock


\bibitem[Liu et~al\mbox{.}(2024a)]%
        {liu2024apigenautomatedpipelinegenerating}
\bibfield{author}{\bibinfo{person}{Zuxin Liu}, \bibinfo{person}{Thai Hoang}, \bibinfo{person}{Jianguo Zhang}, \bibinfo{person}{Ming Zhu}, \bibinfo{person}{Tian Lan}, \bibinfo{person}{Shirley Kokane}, \bibinfo{person}{Juntao Tan}, \bibinfo{person}{Weiran Yao}, \bibinfo{person}{Zhiwei Liu}, \bibinfo{person}{Yihao Feng}, \bibinfo{person}{Rithesh Murthy}, \bibinfo{person}{Liangwei Yang}, \bibinfo{person}{Silvio Savarese}, \bibinfo{person}{Juan~Carlos Niebles}, \bibinfo{person}{Huan Wang}, \bibinfo{person}{Shelby Heinecke}, {and} \bibinfo{person}{Caiming Xiong}.} \bibinfo{year}{2024}\natexlab{a}.
\newblock \bibinfo{title}{APIGen: Automated Pipeline for Generating Verifiable and Diverse Function-Calling Datasets}.
\newblock
\showeprint[arxiv]{2406.18518}~[cs.CL]
\urldef\tempurl%
\url{https://arxiv.org/abs/2406.18518}
\showURL{%
\tempurl}


\bibitem[Lozhkov et~al\mbox{.}(2024)]%
        {lozhkov2024starcoder2stackv2}
\bibfield{author}{\bibinfo{person}{Anton Lozhkov}, \bibinfo{person}{Raymond Li}, \bibinfo{person}{Loubna~Ben Allal}, \bibinfo{person}{Federico Cassano}, {and} \bibinfo{person}{et al.}} \bibinfo{year}{2024}\natexlab{}.
\newblock \bibinfo{title}{StarCoder 2 and The Stack v2: The Next Generation}.
\newblock
\showeprint[arxiv]{2402.19173}~[cs.SE]
\urldef\tempurl%
\url{https://arxiv.org/abs/2402.19173}
\showURL{%
\tempurl}


\bibitem[Lu et~al\mbox{.}(2021)]%
        {lu2021codexglue}
\bibfield{author}{\bibinfo{person}{Shuai Lu}, \bibinfo{person}{Daya Guo}, \bibinfo{person}{Shuo Ren}, \bibinfo{person}{Junjie Huang}, \bibinfo{person}{Alexey Svyatkovskiy}, \bibinfo{person}{Ambrosio Blanco}, \bibinfo{person}{Colin Clement}, \bibinfo{person}{Dawn Drain}, \bibinfo{person}{Daxin Jiang}, \bibinfo{person}{Duyu Tang}, {et~al\mbox{.}}} \bibinfo{year}{2021}\natexlab{}.
\newblock \showarticletitle{Codexglue: A machine learning benchmark dataset for code understanding and generation}.
\newblock \bibinfo{journal}{\emph{arXiv preprint arXiv:2102.04664}} (\bibinfo{year}{2021}).
\newblock


\bibitem[Mu et~al\mbox{.}(2023)]%
        {mu2023developer}
\bibfield{author}{\bibinfo{person}{Fangwen Mu}, \bibinfo{person}{Xiao Chen}, \bibinfo{person}{Lin Shi}, \bibinfo{person}{Song Wang}, {and} \bibinfo{person}{Qing Wang}.} \bibinfo{year}{2023}\natexlab{}.
\newblock \showarticletitle{Developer-intent driven code comment generation}. In \bibinfo{booktitle}{\emph{2023 IEEE/ACM 45th International Conference on Software Engineering (ICSE)}}. IEEE, \bibinfo{pages}{768--780}.
\newblock


\bibitem[Nguyen et~al\mbox{.}(2024)]%
        {vannguyen2024surveysmalllanguagemodels}
\bibfield{author}{\bibinfo{person}{Chien~Van Nguyen}, \bibinfo{person}{Xuan Shen}, \bibinfo{person}{Ryan Aponte}, \bibinfo{person}{Yu Xia}, \bibinfo{person}{Samyadeep Basu}, \bibinfo{person}{Zhengmian Hu}, \bibinfo{person}{Jian Chen}, \bibinfo{person}{Mihir Parmar}, \bibinfo{person}{Sasidhar Kunapuli}, \bibinfo{person}{Joe Barrow}, \bibinfo{person}{Junda Wu}, \bibinfo{person}{Ashish Singh}, \bibinfo{person}{Yu Wang}, \bibinfo{person}{Jiuxiang Gu}, \bibinfo{person}{Franck Dernoncourt}, \bibinfo{person}{Nesreen~K. Ahmed}, \bibinfo{person}{Nedim Lipka}, \bibinfo{person}{Ruiyi Zhang}, \bibinfo{person}{Xiang Chen}, \bibinfo{person}{Tong Yu}, \bibinfo{person}{Sungchul Kim}, \bibinfo{person}{Hanieh Deilamsalehy}, \bibinfo{person}{Namyong Park}, \bibinfo{person}{Mike Rimer}, \bibinfo{person}{Zhehao Zhang}, \bibinfo{person}{Huanrui Yang}, \bibinfo{person}{Ryan~A. Rossi}, {and} \bibinfo{person}{Thien~Huu Nguyen}.} \bibinfo{year}{2024}\natexlab{}.
\newblock \bibinfo{title}{A Survey of Small Language Models}.
\newblock
\showeprint[arxiv]{2410.20011}~[cs.CL]
\urldef\tempurl%
\url{https://arxiv.org/abs/2410.20011}
\showURL{%
\tempurl}


\bibitem[NousResearch(2023)]%
        {Nous-Hermes-13b}
\bibfield{author}{\bibinfo{person}{NousResearch}.} \bibinfo{year}{2023}\natexlab{}.
\newblock \bibinfo{title}{Nous-Hermes-13b}.
\newblock
\urldef\tempurl%
\url{https://huggingface.co/NousResearch/Nous-Hermes-13b}
\showURL{%
\tempurl}


\bibitem[Papineni et~al\mbox{.}(2002)]%
        {papineni2002bleu}
\bibfield{author}{\bibinfo{person}{Kishore Papineni}, \bibinfo{person}{Salim Roukos}, \bibinfo{person}{Todd Ward}, {and} \bibinfo{person}{Wei-Jing Zhu}.} \bibinfo{year}{2002}\natexlab{}.
\newblock \showarticletitle{Bleu: a method for automatic evaluation of machine translation}. In \bibinfo{booktitle}{\emph{Proceedings of the 40th annual meeting of the Association for Computational Linguistics}}. \bibinfo{pages}{311--318}.
\newblock


\bibitem[Parra et~al\mbox{.}(2018)]%
        {parra2018analysis}
\bibfield{author}{\bibinfo{person}{Eugenio Parra}, \bibinfo{person}{Jose~Luis de~la Vara}, {and} \bibinfo{person}{Luis Alonso}.} \bibinfo{year}{2018}\natexlab{}.
\newblock \showarticletitle{Analysis of requirements quality evolution}. In \bibinfo{booktitle}{\emph{Proceedings of the 40th International Conference on Software Engineering: Companion Proceeedings}}. \bibinfo{pages}{199--200}.
\newblock


\bibitem[Patil et~al\mbox{.}(2023)]%
        {patil2023gorilla}
\bibfield{author}{\bibinfo{person}{Shishir~G Patil}, \bibinfo{person}{Tianjun Zhang}, \bibinfo{person}{Xin Wang}, {and} \bibinfo{person}{Joseph~E Gonzalez}.} \bibinfo{year}{2023}\natexlab{}.
\newblock \showarticletitle{Gorilla: Large language model connected with massive apis}.
\newblock \bibinfo{journal}{\emph{arXiv preprint arXiv:2305.15334}} (\bibinfo{year}{2023}).
\newblock


\bibitem[Pinnaparaju et~al\mbox{.}({[n.\,d.]})]%
        {stable-code-3b}
\bibfield{author}{\bibinfo{person}{Nikhil Pinnaparaju}, \bibinfo{person}{Reshinth Adithyan}, \bibinfo{person}{Duy Phung}, \bibinfo{person}{Jonathan Tow}, \bibinfo{person}{James Baicoianu}, {and} \bibinfo{person}{Nathan Cooper}.} \bibinfo{year}{[n.\,d.]}\natexlab{}.
\newblock \bibinfo{title}{Stable Code 3B}.
\newblock
\urldef\tempurl%
\url{[https://huggingface.co/stabilityai/stable-code-3b](https://huggingface.co/stabilityai/stable-code-3b)}
\showURL{%
\tempurl}


\bibitem[Qin et~al\mbox{.}(2023)]%
        {qin2023toolllm}
\bibfield{author}{\bibinfo{person}{Yujia Qin}, \bibinfo{person}{Shihao Liang}, \bibinfo{person}{Yining Ye}, \bibinfo{person}{Kunlun Zhu}, \bibinfo{person}{Lan Yan}, \bibinfo{person}{Yaxi Lu}, \bibinfo{person}{Yankai Lin}, \bibinfo{person}{Xin Cong}, \bibinfo{person}{Xiangru Tang}, \bibinfo{person}{Bill Qian}, {et~al\mbox{.}}} \bibinfo{year}{2023}\natexlab{}.
\newblock \showarticletitle{Toolllm: Facilitating large language models to master 16000+ real-world apis}.
\newblock \bibinfo{journal}{\emph{arXiv preprint arXiv:2307.16789}} (\bibinfo{year}{2023}).
\newblock


\bibitem[Shanahan(2024)]%
        {shanahan2024talking}
\bibfield{author}{\bibinfo{person}{Murray Shanahan}.} \bibinfo{year}{2024}\natexlab{}.
\newblock \showarticletitle{Talking about large language models}.
\newblock \bibinfo{journal}{\emph{Commun. ACM}} \bibinfo{volume}{67}, \bibinfo{number}{2} (\bibinfo{year}{2024}), \bibinfo{pages}{68--79}.
\newblock


\bibitem[Song et~al\mbox{.}(2023)]%
        {song2023comprehensive}
\bibfield{author}{\bibinfo{person}{Yisheng Song}, \bibinfo{person}{Ting Wang}, \bibinfo{person}{Puyu Cai}, \bibinfo{person}{Subrota~K Mondal}, {and} \bibinfo{person}{Jyoti~Prakash Sahoo}.} \bibinfo{year}{2023}\natexlab{}.
\newblock \showarticletitle{A comprehensive survey of few-shot learning: Evolution, applications, challenges, and opportunities}.
\newblock \bibinfo{journal}{\emph{Comput. Surveys}} \bibinfo{volume}{55}, \bibinfo{number}{13s} (\bibinfo{year}{2023}), \bibinfo{pages}{1--40}.
\newblock


\bibitem[Tang et~al\mbox{.}(2023)]%
        {tang2023toolalpaca}
\bibfield{author}{\bibinfo{person}{Qiaoyu Tang}, \bibinfo{person}{Ziliang Deng}, \bibinfo{person}{Hongyu Lin}, \bibinfo{person}{Xianpei Han}, \bibinfo{person}{Qiao Liang}, \bibinfo{person}{Boxi Cao}, {and} \bibinfo{person}{Le Sun}.} \bibinfo{year}{2023}\natexlab{}.
\newblock \showarticletitle{Toolalpaca: Generalized tool learning for language models with 3000 simulated cases}.
\newblock \bibinfo{journal}{\emph{arXiv preprint arXiv:2306.05301}} (\bibinfo{year}{2023}).
\newblock


\bibitem[Tian et~al\mbox{.}(2024)]%
        {tian2024debugbench}
\bibfield{author}{\bibinfo{person}{Runchu Tian}, \bibinfo{person}{Yining Ye}, \bibinfo{person}{Yujia Qin}, \bibinfo{person}{Xin Cong}, \bibinfo{person}{Yankai Lin}, \bibinfo{person}{Yinxu Pan}, \bibinfo{person}{Yesai Wu}, \bibinfo{person}{Haotian Hui}, \bibinfo{person}{Weichuan Liu}, \bibinfo{person}{Zhiyuan Liu}, {et~al\mbox{.}}} \bibinfo{year}{2024}\natexlab{}.
\newblock \showarticletitle{Debugbench: Evaluating debugging capability of large language models}.
\newblock \bibinfo{journal}{\emph{arXiv preprint arXiv:2401.04621}} (\bibinfo{year}{2024}).
\newblock


\bibitem[Wang et~al\mbox{.}(2024b)]%
        {wang2024comprehensivesurveysmalllanguage}
\bibfield{author}{\bibinfo{person}{Fali Wang}, \bibinfo{person}{Zhiwei Zhang}, \bibinfo{person}{Xianren Zhang}, \bibinfo{person}{Zongyu Wu}, \bibinfo{person}{Tzuhao Mo}, \bibinfo{person}{Qiuhao Lu}, \bibinfo{person}{Wanjing Wang}, \bibinfo{person}{Rui Li}, \bibinfo{person}{Junjie Xu}, \bibinfo{person}{Xianfeng Tang}, \bibinfo{person}{Qi He}, \bibinfo{person}{Yao Ma}, \bibinfo{person}{Ming Huang}, {and} \bibinfo{person}{Suhang Wang}.} \bibinfo{year}{2024}\natexlab{b}.
\newblock \bibinfo{title}{A Comprehensive Survey of Small Language Models in the Era of Large Language Models: Techniques, Enhancements, Applications, Collaboration with LLMs, and Trustworthiness}.
\newblock
\showeprint[arxiv]{2411.03350}~[cs.CL]
\urldef\tempurl%
\url{https://arxiv.org/abs/2411.03350}
\showURL{%
\tempurl}


\bibitem[Wang et~al\mbox{.}(2024a)]%
        {wang2024deepedit}
\bibfield{author}{\bibinfo{person}{Yiwei Wang}, \bibinfo{person}{Muhao Chen}, \bibinfo{person}{Nanyun Peng}, {and} \bibinfo{person}{Kai-Wei Chang}.} \bibinfo{year}{2024}\natexlab{a}.
\newblock \bibinfo{title}{DeepEdit: Knowledge Editing as Decoding with Constraints}.
\newblock
\showeprint[arxiv]{2401.10471}~[cs.CL]


\bibitem[Wang et~al\mbox{.}(2020)]%
        {wang2020generalizing}
\bibfield{author}{\bibinfo{person}{Yaqing Wang}, \bibinfo{person}{Quanming Yao}, \bibinfo{person}{James~T Kwok}, {and} \bibinfo{person}{Lionel~M Ni}.} \bibinfo{year}{2020}\natexlab{}.
\newblock \showarticletitle{Generalizing from a few examples: A survey on few-shot learning}.
\newblock \bibinfo{journal}{\emph{ACM computing surveys (csur)}} \bibinfo{volume}{53}, \bibinfo{number}{3} (\bibinfo{year}{2020}), \bibinfo{pages}{1--34}.
\newblock


\bibitem[Willison(2023)]%
        {fake-completion}
\bibfield{author}{\bibinfo{person}{Simon Willison}.} \bibinfo{year}{2023}\natexlab{}.
\newblock \bibinfo{title}{Delimiters won’t save you from prompt injection}.
\newblock
\urldef\tempurl%
\url{https://simonwillison.net/2023/May/11/delimiters-wont-save-you/}
\showURL{%
\tempurl}


\bibitem[Wohlin et~al\mbox{.}(2012)]%
        {wohlin2012experimentation}
\bibfield{author}{\bibinfo{person}{Claes Wohlin}, \bibinfo{person}{Per Runeson}, \bibinfo{person}{Martin H{\"o}st}, \bibinfo{person}{Magnus~C Ohlsson}, \bibinfo{person}{Bj{\"o}rn Regnell}, \bibinfo{person}{Anders Wessl{\'e}n}, {et~al\mbox{.}}} \bibinfo{year}{2012}\natexlab{}.
\newblock \bibinfo{booktitle}{\emph{Experimentation in software engineering}}. Vol.~\bibinfo{volume}{236}.
\newblock \bibinfo{publisher}{Springer}.
\newblock


\bibitem[Wu et~al\mbox{.}(2024)]%
        {wu2024newerallmsecurity}
\bibfield{author}{\bibinfo{person}{Fangzhou Wu}, \bibinfo{person}{Ning Zhang}, \bibinfo{person}{Somesh Jha}, \bibinfo{person}{Patrick McDaniel}, {and} \bibinfo{person}{Chaowei Xiao}.} \bibinfo{year}{2024}\natexlab{}.
\newblock \bibinfo{title}{A New Era in LLM Security: Exploring Security Concerns in Real-World LLM-based Systems}.
\newblock
\showeprint[arxiv]{2402.18649}~[cs.CR]
\urldef\tempurl%
\url{https://arxiv.org/abs/2402.18649}
\showURL{%
\tempurl}


\bibitem[Yan et~al\mbox{.}(2024a)]%
        {yan2024protecting}
\bibfield{author}{\bibinfo{person}{Biwei Yan}, \bibinfo{person}{Kun Li}, \bibinfo{person}{Minghui Xu}, \bibinfo{person}{Yueyan Dong}, \bibinfo{person}{Yue Zhang}, \bibinfo{person}{Zhaochun Ren}, {and} \bibinfo{person}{Xiuzhen Cheng}.} \bibinfo{year}{2024}\natexlab{a}.
\newblock \showarticletitle{On protecting the data privacy of large language models (llms): A survey}.
\newblock \bibinfo{journal}{\emph{arXiv preprint arXiv:2403.05156}} (\bibinfo{year}{2024}).
\newblock


\bibitem[Yan et~al\mbox{.}(2024b)]%
        {berkeley-function-calling-leaderboard}
\bibfield{author}{\bibinfo{person}{Fanjia Yan}, \bibinfo{person}{Huanzhi Mao}, \bibinfo{person}{Charlie Cheng-Jie Ji}, \bibinfo{person}{Tianjun Zhang}, \bibinfo{person}{Shishir~G. Patil}, \bibinfo{person}{Ion Stoica}, {and} \bibinfo{person}{Joseph~E. Gonzalez}.} \bibinfo{year}{2024}\natexlab{b}.
\newblock \showarticletitle{Berkeley Function Calling Leaderboard}. \bibinfo{howpublished}{\url{https://gorilla.cs.berkeley.edu/blogs/8_berkeley_function_calling_leaderboard.html}}.
\newblock


\bibitem[Yao et~al\mbox{.}(2022)]%
        {yao2022react}
\bibfield{author}{\bibinfo{person}{Shunyu Yao}, \bibinfo{person}{Jeffrey Zhao}, \bibinfo{person}{Dian Yu}, \bibinfo{person}{Nan Du}, \bibinfo{person}{Izhak Shafran}, \bibinfo{person}{Karthik Narasimhan}, {and} \bibinfo{person}{Yuan Cao}.} \bibinfo{year}{2022}\natexlab{}.
\newblock \showarticletitle{React: Synergizing reasoning and acting in language models}.
\newblock \bibinfo{journal}{\emph{arXiv preprint arXiv:2210.03629}} (\bibinfo{year}{2022}).
\newblock


\bibitem[Zhang et~al\mbox{.}(2023)]%
        {zhang2023toolcoder}
\bibfield{author}{\bibinfo{person}{Kechi Zhang}, \bibinfo{person}{Huangzhao Zhang}, \bibinfo{person}{Ge Li}, \bibinfo{person}{Jia Li}, \bibinfo{person}{Zhuo Li}, {and} \bibinfo{person}{Zhi Jin}.} \bibinfo{year}{2023}\natexlab{}.
\newblock \showarticletitle{Toolcoder: Teach code generation models to use api search tools}.
\newblock \bibinfo{journal}{\emph{arXiv preprint arXiv:2305.04032}} (\bibinfo{year}{2023}).
\newblock


\bibitem[Zhao et~al\mbox{.}(2023)]%
        {zhao2023survey}
\bibfield{author}{\bibinfo{person}{Wayne~Xin Zhao}, \bibinfo{person}{Kun Zhou}, \bibinfo{person}{Junyi Li}, \bibinfo{person}{Tianyi Tang}, \bibinfo{person}{Xiaolei Wang}, \bibinfo{person}{Yupeng Hou}, \bibinfo{person}{Yingqian Min}, \bibinfo{person}{Beichen Zhang}, \bibinfo{person}{Junjie Zhang}, \bibinfo{person}{Zican Dong}, {et~al\mbox{.}}} \bibinfo{year}{2023}\natexlab{}.
\newblock \showarticletitle{A survey of large language models}.
\newblock \bibinfo{journal}{\emph{arXiv preprint arXiv:2303.18223}} (\bibinfo{year}{2023}).
\newblock


\bibitem[Zheng et~al\mbox{.}(2024)]%
        {zheng2024reviewedgelargelanguage}
\bibfield{author}{\bibinfo{person}{Yue Zheng}, \bibinfo{person}{Yuhao Chen}, \bibinfo{person}{Bin Qian}, \bibinfo{person}{Xiufang Shi}, \bibinfo{person}{Yuanchao Shu}, {and} \bibinfo{person}{Jiming Chen}.} \bibinfo{year}{2024}\natexlab{}.
\newblock \bibinfo{title}{A Review on Edge Large Language Models: Design, Execution, and Applications}.
\newblock
\showeprint[arxiv]{2410.11845}~[cs.DC]
\urldef\tempurl%
\url{https://arxiv.org/abs/2410.11845}
\showURL{%
\tempurl}


\bibitem[Zhong et~al\mbox{.}(2025)]%
        {zhong2025complexfuncbench}
\bibfield{author}{\bibinfo{person}{Lucen Zhong}, \bibinfo{person}{Zhengxiao Du}, \bibinfo{person}{Xiaohan Zhang}, \bibinfo{person}{Haiyi Hu}, {and} \bibinfo{person}{Jie Tang}.} \bibinfo{year}{2025}\natexlab{}.
\newblock \showarticletitle{ComplexFuncBench: Exploring Multi-Step and Constrained Function Calling under Long-Context Scenario}.
\newblock \bibinfo{journal}{\emph{arXiv preprint arXiv:2501.10132}} (\bibinfo{year}{2025}).
\newblock


\bibitem[Zhu et~al\mbox{.}(2024)]%
        {zhu2024survey}
\bibfield{author}{\bibinfo{person}{Xunyu Zhu}, \bibinfo{person}{Jian Li}, \bibinfo{person}{Yong Liu}, \bibinfo{person}{Can Ma}, {and} \bibinfo{person}{Weiping Wang}.} \bibinfo{year}{2024}\natexlab{}.
\newblock \showarticletitle{A survey on model compression for large language models}.
\newblock \bibinfo{journal}{\emph{Transactions of the Association for Computational Linguistics}}  \bibinfo{volume}{12} (\bibinfo{year}{2024}), \bibinfo{pages}{1556--1577}.
\newblock


\end{thebibliography}
\end{document}